\documentclass{article}

\usepackage[table]{xcolor}  
\usepackage{graphicx}
\usepackage{pifont}
\usepackage{amsmath}
\usepackage{xcolor}
\usepackage{comment}
\usepackage{pifont}  
\usepackage[most]{tcolorbox} 
\tcbuselibrary{breakable}
\usepackage{booktabs}
\usepackage[table]{xcolor}
\usepackage{tabularx}
\usepackage[preprint]{neurips_2026}

\usepackage[utf8]{inputenc} 
\usepackage[T1]{fontenc}    
\usepackage{hyperref}       
\usepackage{url}            
\usepackage{booktabs}       
\usepackage{amsfonts}       
\usepackage{nicefrac}       
\usepackage{microtype}      
\usepackage{xcolor}         

\title{SpatialForge: Bootstrapping 3D-Aware Spatial Reasoning from Open-World 2D Images}

\author{
\textbf{Zishan Liu}$^{1,2}$ \quad
\textbf{Ruoxi Zang}$^{2}$ \quad
\textbf{Yanglin Zhang}$^{2}$ \quad
\textbf{Wei Liu}$^{2}$ \quad
\textbf{Yin Zhang}$^{2}$ \\
\textbf{Jian Yao}$^{2}$ \quad
\textbf{Jiayin Zheng}$^{2}$\thanks{Corresponding author.} \quad
\textbf{Zhengzhe Liu}$^{1}$\footnotemark[1] \\
$^{1}$Lingnan University \qquad
$^{2}$XPENG Robotics \\
}

\usepackage{booktabs} 
\usepackage{multirow} 
\usepackage{makecell} 
\usepackage{wrapfig}

\usepackage{caption}
\begin{document}
\setcitestyle{numbers, square}

\maketitle

\begin{abstract}
Recent advancements in Large Vision-Language Models (VLMs) have demonstrated exceptional semantic understanding, yet these models consistently struggle with spatial reasoning, often failing at fundamental geometric tasks such as depth ordering and precise coordinate grounding. 
Recent efforts introduce spatial supervision from scene-centric datasets (e.g., multi-view scans or indoor video), but are constrained by the limited number of underlying scenes. As a result, the scale and diversity of such data remain significantly smaller than those of web-scale 2D image collections.
To address this limitation, we propose SpatialForge, a scalable data synthesis pipeline that transforms in-the-wild 2D images into spatial reasoning supervision. 
Our approach decomposes spatial reasoning into perception and relation, and constructs structured supervision signals covering depth, layout, and viewpoint-dependent reasoning, with automatic verification to ensure data quality. 
Based on this pipeline, we build SpatialForge-10M, a large-scale dataset containing 10 million spatial QA pairs. Extensive experiments across multiple spatial reasoning benchmarks demonstrate that training on SpatialForge-10M significantly improves the spatial reasoning ability of standard VLMs, highlighting the effectiveness of scaling 2D data for 3D-aware spatial reasoning.
\end{abstract}

\section{Introduction}

Large Vision-Language Models (VLMs) have achieved remarkable success in aligning visual inputs with human semantics, demonstrating strong capabilities in tasks ranging from complex scene comprehension to visual question answering~\citep{liu2023visual, team2023gemini, yang2025qwen3}. However, despite this strong semantic understanding, contemporary VLMs exhibit a notable limitation: they often struggle with spatial reasoning. While these models can accurately identify objects and retrieve semantic facts, they frequently encounter difficulties with fundamental geometric tasks, such as determining fine-grained depth ordering (e.g., near/far relationships), grounding precise spatial region descriptions, or inferring layouts from alternative viewpoints~\citep{tong2024eyes, yang2025thinking}. This discrepancy arises in part because VLMs are predominantly trained on web-scale image-text pairs that emphasize object-centric semantics rather than topological and geometric relationships. Addressing this spatial deficit is essential for deploying VLMs in physically grounded systems, including embodied robotics~\citep{zitkovich2023rt}, autonomous navigation~\citep{tian2024drivevlm}, and augmented reality~\citep{chandrasegaran2024hourvideo}.

\begin{table}[t]
\centering
\caption{\textbf{Comparison of SpatialForge with existing spatial reasoning datasets.}}
\label{tab:dataset_compare}
\begin{tabular}{lccccc}
\toprule
Dataset & \# Scenes & \# Source Data  & \# Spatial QAs & Scenario \\
\midrule
Spatial-MLLM~\citep{wu2025spatial} & 1.5k & 1.5k videos & 120k & Indoor  \\
SpatialLadder~\citep{li2025spatialladder} & $\sim$20k & 11k images, 9k videos  & 26k & Indoor \\
SPAR-7M~\citep{zhang2025flatland} & 4k & 4k videos  & 7M & Indoor \\
SpatialQA~\citep{cai2025spatialbot} & $\sim$723k & 723k images  & 0.9M & Embodied  \\
\midrule
SpatialForge & $\sim$2M & 2M images & 10M & Open-world \\
\bottomrule
\end{tabular}

\end{table}

The primary bottleneck in endowing VLMs with spatial awareness is the scarcity of scalable, high-quality spatial supervision. Recent efforts to address this bottleneck have largely diverged into two trajectories. The first introduces explicit 3D representations into the VLM architecture~\citep{hong20233d, qi2024shapellm, zhu2025llava3dsimpleeffectivepathway}. While effective, this approach typically requires specialized 3D encoders, modifies the unified architecture of existing VLMs, and relies on multi-modal sensor inputs that may not be available in unconstrained, real-world deployments. 
The second trajectory attempts to synthesize spatial question–answer (QA) pairs from existing scene-centric datasets~\citep{li2025spatialladder, zhang2025flatland, deng2025internspatial}. In practice, these approaches typically rely on a limited set of indoor environments (e.g., ScanNet~\cite{dai2017scannet} and ScanNet++~\cite{yeshwanth2023scannet++}), which are further expanded through multi-view images or video frames to construct training data. 
While effective in introducing geometric supervision, this paradigm is inherently constrained by the number of underlying scenes. As a result, although many visual samples can be generated, their diversity remains limited by the original environments. This leads to two key limitations: (1) \textbf{limited scale}, as acquiring new scenes and annotating fine-grained spatial relationships across objects is expensive, and (2) \textbf{limited diversity}, since samples are repeatedly drawn from similar layouts and object configurations. Consequently, these datasets often exhibit domain bias toward indoor settings and may struggle to generalize to open-world scenarios.

To overcome the limitations of scene-centric data and the difficulty of scaling spatial supervision, we explore an alternative approach: extracting structured spatial signals directly from large-scale in-the-wild 2D images. Unlike prior approaches that rely on a limited set of scenes, our data is drawn from diverse open-world images, where each image effectively introduces a new scene, leading to substantially greater scene diversity. In this paper, we introduce \textbf{SpatialForge}, an automated and scalable data synthesis engine that transforms single-view 2D images into structured spatial reasoning data, without relying on 3D scene data. 
into two hierarchical cognitive levels: \textbf{spatial perception} and \textbf{spatial relation}. The spatial perception level focuses on precise visual grounding, referring, and counting, aiming to accurately localize and describe objects based on direct visual evidence.
At the spatial relation level, to further bridge the gap between 2D observations and 3D spatial understanding, we focus on two fundamental aspects: depth and directional relations. For depth, we construct supervision signals that emphasize relative distance and occlusion, encouraging the model to infer near–far relationships beyond 2D appearance cues. For directional reasoning, we go beyond generic left–right relations by further introducing perspective-dependent ones. We augment training with perspective-aware transformations by introducing human-centric viewpoints and synthesizing QA pairs that require reasoning from alternative perspectives. 
Together, these designs promote a more robust and 3D-consistent understanding of spatial relationships.


Leveraging this automated pipeline, we construct \textbf{SpatialForge-10M}, a large-scale, open-world dataset containing 10 million spatial QA pairs derived from 2 million curated images. 
SpatialForge-10M spans diverse environments, features an open-vocabulary category space, and covers a comprehensive taxonomy of spatial tasks. 
%
Extensive experiments demonstrate the effectiveness of this data-centric approach. 
By fine-tuning standard VLMs on SpatialForge, we observe substantial improvements in spatial reasoning performance across multiple benchmarks, indicating that large-scale 2D-derived supervision can effectively enhance spatial-aware reasoning without requiring additional 3D-specific inputs or architectural modifications. The dataset will be released upon publication.


Our key contributions are summarized as follows:
\begin{itemize}
\item We propose \textbf{SpatialForge}, a scalable, automated data synthesis engine that extracts structured, \textbf{3D-aware spatial supervision} from single-view 2D images, effectively mitigating the scalability limits of explicit 3D annotations.
\item We construct \textbf{SpatialForge-10M}, a large-scale open-world spatial QA dataset that can improve spatial reasoning through two complementary subcategories: spatial perception and spatial relations, covering 6 spatial tasks.
\item We demonstrate through extensive experiments that data-centric scaling via SpatialForge effectively enhances spatial reasoning in standard VLMs, achieving state-of-the-art performance across multiple benchmarks without requiring architectural modifications.
\end{itemize}

\begin{figure}[t]
  \centering
  \includegraphics[width=0.95\columnwidth]{ 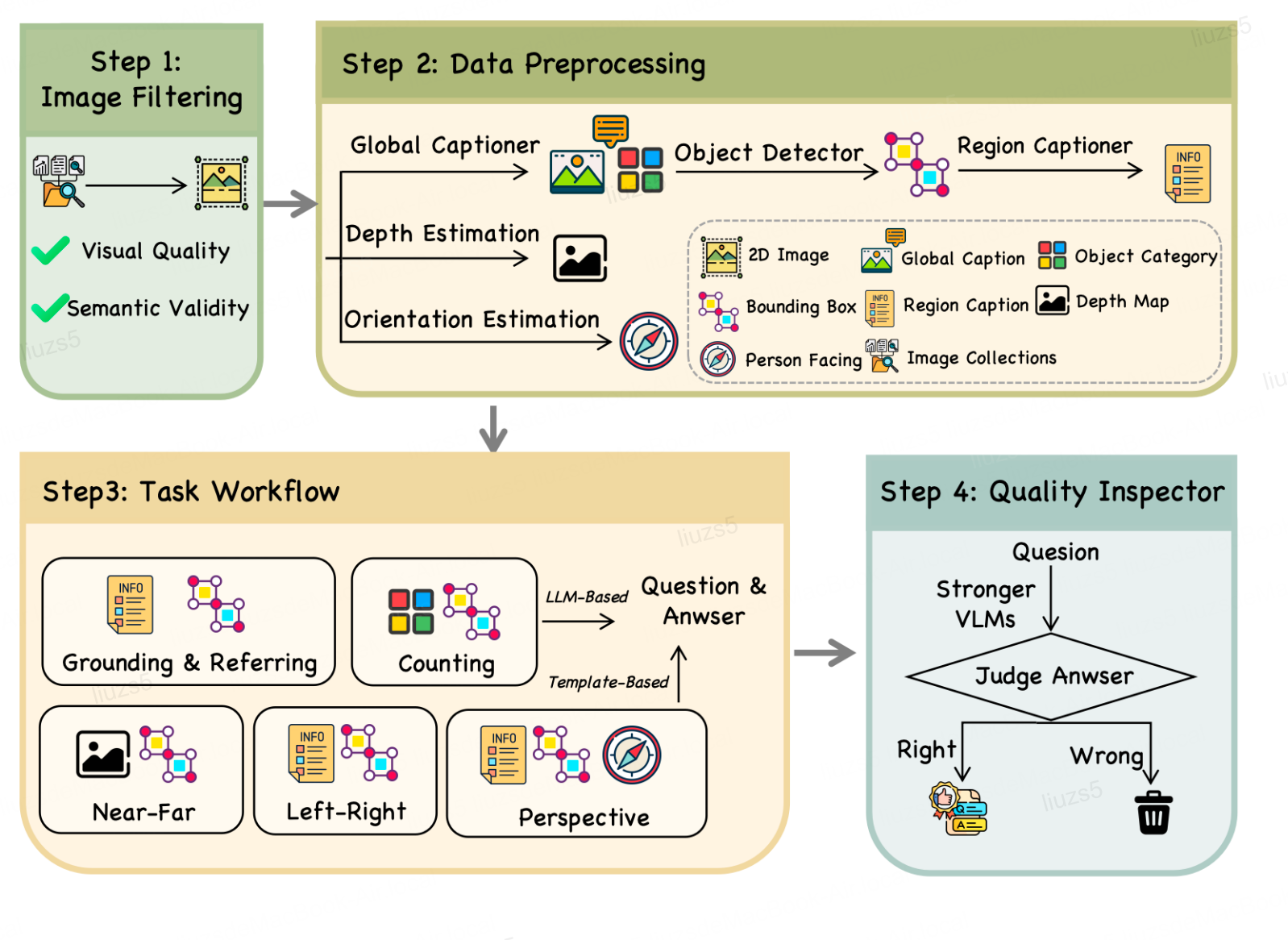}
  \caption{\textbf{Overview of the SpatialForge pipeline.} Our pipeline consists of four steps: filtering images, extracting object-level information, generating spatial QA tasks, and verifying quality.}
  \label{fig:data engine pipeline}
\end{figure}

\section{Related Work}
\label{sec:related_work}

\subsection{Spatial Reasoning Paradigms in VLMs}
While modern vision-language models (VLMs) achieve strong performance across broad multimodal benchmarks~\citep{liu2024mmbench,mathew2021docvqa,fu2024ocrbench,yue2024mmmu,li2024llava,achiam2023gpt}, they continue to face challenges with geometrically grounded tasks, such as depth estimation, viewpoint transformations, and multi-step spatial logic~\citep{yang2025thinking,jia2025omnispatial,yin2025spatial}. Efforts to mitigate these spatial limitations generally follow three paradigms. 
The first introduces \textit{explicit 3D modalities} (e.g., point clouds, voxel grids, or metric depth maps) directly into the VLM architecture~\citep{hong20233d,qi2024shapellm,zhu2025llava3dsimpleeffectivepathway,xu2024pointllm,cheng2024spatialrgpt}. While effective in controlled settings, these methods require specialized 3D encoders and multi-modal inputs that may not be available in unconstrained open-world deployments. 
The second paradigm focuses on \textit{structured intermediate representations}, prompting VLMs to generate textual spatial graphs, coordinate traces, or chain-of-thought rationales~\citep{chen2024spatialvlm,yin2025spatial,ouyang2025spacer,liu2025spatialcot,li2025imagine}. Though they improve compositional logic, these methods often rely on curated task-specific formats that can be challenging to generalize in the wild. 
The third paradigm employs \textit{inference-time tool augmentation}, using external depth estimators or robotic perception modules to provide auxiliary geometric evidence~\citep{tian2026last,suris2023vipergpt,driess2023palm}. This adds latency and computational overhead. In contrast, we adopt a purely data-centric approach: we improve the spatial capability of standard VLMs through large-scale fine-tuning, without requiring architectural changes, multi-modal sensor inputs, or external inference tools.

\subsection{Spatial Supervision and Datasets}
A central bottleneck in training spatially aware VLMs is the lack of scalable and diverse geometric supervision. 
Early spatial datasets primarily focus on 2D relationships, such as region grounding, counting, and bounding-box localization~\citep{zhang2024ferret,ranasinghe2024learning,chen2023shikra,wang2024picture}. 
While effective for visual alignment, these datasets provide limited supervision for depth, layout, and viewpoint-dependent reasoning.
To address this limitation, recent works construct spatial reasoning data from scene-centric sources~\citep{li2025spatialladder,deng2025internspatial,zhang2025flatland,song2025robospatial,li2024proximity}. 
In practice, these approaches typically generate images or videos from a limited set of underlying scenes and synthesize spatial QA pairs accordingly. 
Although this process introduces stronger geometric signals, it is inherently constrained by the cost of acquiring new scenes and annotating fine-grained spatial relationships. 
As summarized in Table~\ref{tab:dataset_compare}, existing spatial datasets are generally limited in both scale and diversity, as many samples are derived from the same environments and are predominantly restricted to indoor or structured settings. 
On the contrary, we leverage large-scale open-world images and propose an automatic data synthesis pipeline and construct a comprehensive supervision signal that encompasses both fundamental spatial perception and high-order allocentric relations. This enables the model to not only localize objects in 3D space but also internalize the underlying geometric logic required for open-world spatial intelligence, all without relying on constrained 3D annotations. Our method provides a comprehensive and scalable solution for enhancing the 3D-aware reasoning capabilities of multimodal models in the wild.

\begin{figure}[t]
  \centering
  \includegraphics[width=\columnwidth]{ 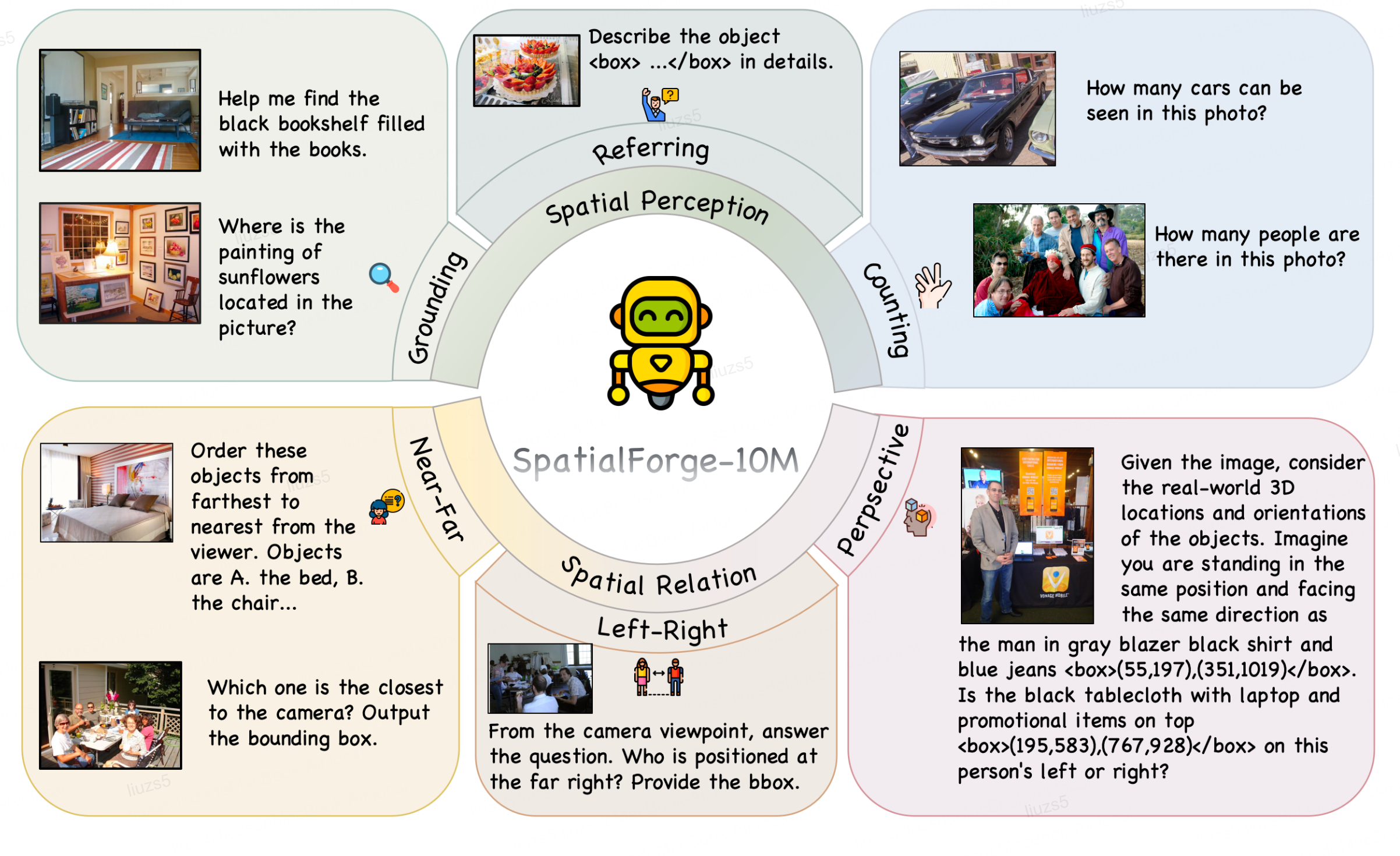}
  \caption{
\textbf{Overview of SpatialForge-10M.} The dataset covers six tasks to improve sptial reasoning capability from 2D images. 
}
  \label{fig:SpatialForge_overview}
\end{figure}

\section{Methodology}
\label{sec:method}

\subsection{Overview}
Our work aims to enhance the spatial reasoning capabilities of VLMs through large-scale synthetic data. To overcome the limitations of existing datasets in scale and open-world diversity, we introduce a scalable, fully automated data synthesis engine that converts 2D images into structured 3D spatial supervision. As illustrated in Figure~\ref{fig:data engine pipeline}, our data synthesis pipeline starts with a single image and progressively generates spatial reasoning data step by step by integrating specialized models.  
We provide a detailed formalization of each step of our data synthesis pipeline in Sec.~\ref{data engine}, followed by dataset statistics and analysis in Sec.~\ref{statistics}.

\subsection{Task Taxonomy}
\label{sec:task_taxonomy}

As shown in Figure~\ref{fig:SpatialForge_overview}, SpatialForge-10M covers six task families organized into two levels: 
\textbf{spatial perception} and \textbf{spatial relation}. 
Spatial perception focuses on extracting object-level information that is directly observable from the image, including localization, description, and counting. These tasks rely primarily on visual evidence and establish reliable grounding between objects and their representations.
In contrast, spatial relation focuses on reasoning about relationships between objects based on geometric cues, such as relative depth and horizontal layout. These relations are often ambiguous under 2D projection and require the model to infer underlying spatial structure beyond the image plane.
These two levels are complementary: spatial perception provides reliable object-level grounding, which serves as the basis for spatial relation reasoning that requires comparing objects and resolving ambiguities beyond direct visual evidence. 
In particular, accurate perception is critical for establishing consistent object references, which directly affects the reliability of downstream spatial reasoning.

\textbf{Spatial Perception.}
Spatial perception tasks focus on extracting object-level information that is directly observable from the image, such as localization and identification.
   (i) \textit{Grounding.} 
    Given a description, the model localizes the corresponding object or region.
    (ii) \textit{Referring.} 
    Given a region, the model identifies or describes the object within it.
    (iii) \textit{Counting.} 
Given a query, the model counts objects that satisfy the condition.

\textbf{Spatial Relation.}
Spatial relation tasks focus on reasoning about relationships between objects, often requiring resolving ambiguities beyond direct visual evidence.
(i) \textit{Near--Far.} 
    The model determines relative depth ordering between objects.
(ii) \textit{Left--Right.} 
The model identifies objects by their horizontal relation to a reference object.
(iii) \textit{Perspective.} 
The model interprets spatial relations under a specified viewpoint.

\subsection{SpatialForge Pipeline}
\label{data engine}

\subsubsection{Image Filtering}

As shown in Figure~\ref{fig:data engine pipeline} step 1, 
we first filter the raw image pool to ensure both visual quality and physical realism. 
At the visual level, we remove low-quality images such as those that are blurred, poorly exposed, or severely distorted, as these can degrade geometric consistency.
At the semantic level, we use CLIP~\citep{radford2021learning} to distinguish real-world scenes from synthetic or non-physical content. Specifically, we compare image embeddings with a small set of textual anchors (e.g., ``natural scene'' vs. ``GUI interface'') and discard images that are more similar to non-physical categories, such as screenshots or text-heavy documents.
This filtering step ensures that the remaining data provides clean and physically grounded inputs for subsequent spatial reasoning.

\subsubsection{Data Preprocessing}
As shown in Figure~\ref{fig:data engine pipeline} step 2, to transform raw images into structured spatial supervision, we design a multi-stage preprocessing pipeline that integrates multiple expert models, including VLMs~\citep{yang2025qwen3}, open-vocabulary detectors~\citep{liu2024grounding}, depth estimators~\citep{yang2024depth}, and orientation predictors.

\paragraph{Global\&Region Caption.} Given an input image $I$, we first employ a high-capacity VLM as a \textit{global captioner} to produce a holistic scene description $C_{\text{global}}$. A lightweight semantic parsing module then extracts a set of object-centric queries $\mathcal{Q} = \{q_1, q_2, \dots, q_N\}$ from the caption, corresponding to salient entities in the scene.
Each query $q_i$ is fed into an open-vocabulary detector to localize the object, yielding a bounding box $B_i$. Conditioned on each detected region $(B_i, q_i)$, the VLM further generates a fine-grained \textit{region caption} $c_i$, describing both visual attributes and local spatial context.
This process produces a structured object-level representation:
\begin{equation}
\mathcal{O} = \{ o_i \}_{i=1}^N, \quad \text{where } o_i = (B_i, c_i)
\end{equation}

Notably, the region captions are designed to be spatially grounded, as they explicitly encode positional cues (e.g., relative location, foreground/background). This alignment between language and geometry provides direct supervision for spatial perception tasks such as \textit{grounding}, \textit{referring}, and \textit{counting}.

\paragraph{Depth Estimation.}
To capture the underlying 3D structure from a single image, we employ a monocular depth estimator to predict a dense depth map $D \in \mathbb{R}^{H \times W}$.
For each object $o_i = (B_i, c_i)$, we associate it with a depth value by aggregating depth statistics within its bounding box $B_i$. In practice, we compute robust statistics such as the median depth to represent the object's overall position, which mitigates noise from monocular predictions. These object-level depth cues serve as the foundation for constructing near--far relationships in subsequent stages.

\paragraph{Human Orientation Estimation.}
To support viewpoint-dependent spatial reasoning, we additionally estimate the orientation of human subjects in the scene. When a person is detected, we apply an orientation predictor to classify their facing direction relative to the camera.
Given the inherent ambiguity of fine-grained orientation estimation, we adopt a simplified yet reliable formulation by categorizing each person into two canonical states: \textit{facing toward} or \textit{facing away} from the camera. This classification can provide the information to determine whether a viewpoint transformation (e.g., left--right reversal) should be applied.
The resulting orientation signal is attached to the corresponding human object and later used to derive perspective-taking spatial relationships. Please refer to the Appendix~\ref{appendix:data_preprocessing} for more details.

\subsubsection{Task Workflow}

As shown in Figure~\ref{fig:data engine pipeline} step 3, Building upon the structured spatial representation constructed in the preprocessing stage, we design a unified workflow to generate diverse spatial reasoning tasks in a scalable manner. 

\paragraph{Spatial Perception Tasks.}
Tasks such as \emph{grounding}, \emph{referring}, and \emph{counting} are directly constructed from object-level annotations.
\textit{Grounding} is obtained by mapping region captions to their corresponding bounding boxes, forming text-to-region pairs. 
\textit{Referring} reverses this process by using a region as input and predicting its associated object description. 
\textit{Counting} is constructed by aggregating object instances that share the same category or satisfy a given attribute.
These tasks rely on explicit visual evidence and establish reliable object-level grounding. Such grounding is essential for spatial reasoning, which requires comparing multiple objects and resolving relationships beyond direct visual cues.

\paragraph{Spatial Reasoning Tasks.}
Tasks such as \emph{near--far}, \emph{left--right}, and \emph{perspective} are constructed by deriving geometric relationships between objects.

\textit{Near--far} is obtained by estimating object-level depth using a monocular depth predictor and aggregating depth values within each bounding box. We compute complementary statistics (e.g., median and high-percentile depth) and determine pairwise depth ordering based on their agreement, discarding cases with inconsistent depth cues.

\textit{Left--right} is constructed from horizontal spatial arrangements under the camera-centric frame. For each object pair, we determine their ordering based on bounding box geometry, and filter out cases with significant overlap or ambiguous layouts to ensure reliable supervision.

\textit{Perspective} extends left--right relations to a human-centric reference frame. When a human subject is detected, we estimate its orientation (facing toward or away from the camera) and transform spatial relations accordingly, enabling viewpoint-dependent annotations.

These tasks require reasoning over relationships between multiple objects and resolving ambiguities caused by projection and viewpoint, thereby encouraging the model to develop spatial reasoning beyond direct 2D observations. Please refer to the Appendix~\ref{appendix:task_workflow} for more details.

\subsubsection{Quality Inspector}
Multi-stage generation pipelines are inherently prone to error accumulation, which can introduce noisy or inconsistent supervision signals. To mitigate this, we implement a Quality Inspector stage to ensure the precision and reliability of the synthesized samples, as shown in Figure~\ref{fig:data engine pipeline} step 4. We employ a stronger VLM~\citep{yang2025qwen3} as an independent judge to validate each generated question-answer pair. 
For each synthesized sample, we feed the image and question into the inspector and compare its predicted answer against the original generated answer. Only samples with consistent answers pass the inspection and are retained; the rest are discarded. This filtering suppresses error propagation and ensures dataset quality. More details are present in Appendix~\ref{appendix:quality}.

\begin{figure}[t]
  \centering
  \includegraphics[width=0.9\columnwidth]{ 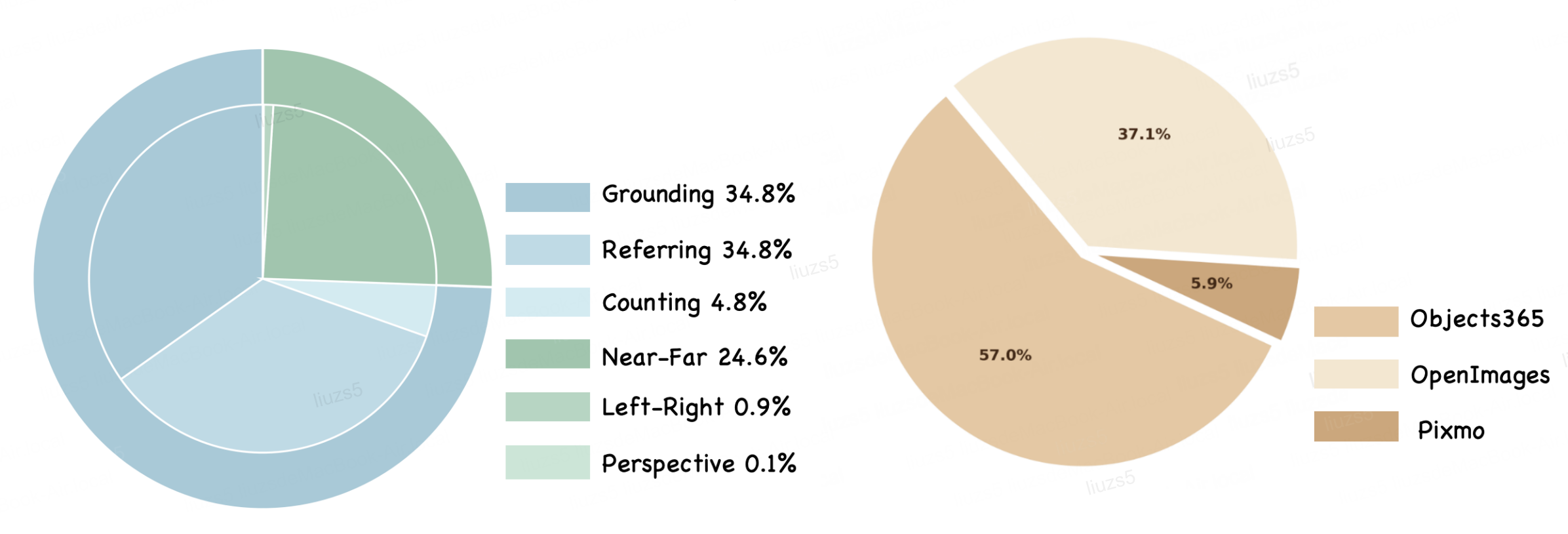}
\caption{\textbf{Distribution of task categories (Left) and data sources (Right) in SpatialForge-10M.}}
    \label{fig:SpatialForge_stats}
\end{figure}


\subsection{Dataset Construction and Statistics}
\label{statistics}

Leveraging our automated data synthesis pipeline, we construct \textbf{SpatialForge-10M}, a large-scale spatial reasoning dataset spanning diverse open-world imagery. As shown in Figure~\ref{fig:SpatialForge_stats}, we aggregate raw images from three primary data sources: Objects365~\citep{shao2019objects365}, OpenImages~\citep{kuznetsova2020open}, and Pixmo~\citep{deitke2025molmo}, ensuring broad coverage across diverse scenes. 
Our pipeline leverages these repositories for their visual content, bypassing original annotations to avoid fixed taxonomies, and distills high-fidelity spatial knowledge, including precise coordinates and spatial relationships. In total, SpatialForge-10M consists of over 2.8 million high-quality images and 10.2 million verified question-answer pairs across six spatial tasks.
Please refer to the Appendix~\ref{sec:appendix_taxonomy} for further statistics.

\section{Experiments}
\label{sec:experiments}
We begin in Section~\ref{sec:experiment_setup} by introducing the baseline models and outlining the specific evaluation benchmarks used. Section~\ref{evaluation_spatial} presents the performance comparison on existing spatial benchmarks to assess the our model's spatial reasoning. Section~\ref{sec:ablation} provides an ablation study to analyze the impact of our data synthesis components on the final performance.

\subsection{Experimental Setup}
\label{sec:experiment_setup}
\paragraph{Baseline.} We adopt Qwen3-VL-2B-Instruct~\citep{yang2025qwen3} as the base model and perform full-parameter supervised fine-tuning using SpatialForge-10M. To maintain the model’s instruction-following capabilities, we incorporate a subset of LLaVA-OneVision-1.5-Instruct-Data~\citep{li2024llava}. For brevity, we refer to this model as "Qwen3-VL-2B" throughout the paper. Detailed training configurations are provided in Appendix~\ref{appendix:appendix_training} .

\begin{table}[t]
\caption{\textbf{Benchmark coverage of spatial perception and relation capabilities.}}
\centering
\resizebox{\linewidth}{!}{  
\small
\setlength{\tabcolsep}{4pt}
\renewcommand{\arraystretch}{1.15}
\begin{tabular}{lccc}
\toprule
\textbf{Benchmark} 
& \textbf{Main Tasks} 
& \textbf{Perception} 
& \textbf{Relation} \\
\midrule
CV-Bench (2D \& 3D)~\citep{tong2024cambrian}
& Counting, positional relations, and distance comparison
& \checkmark 
& \checkmark \\
SPAR~\citep{zhang2025flatland} 
& Distance prediction, spatial relations, and spatial imagination
& \checkmark 
& \checkmark \\
SpaCE10~\citep{Gong2025SpaCE10AC} 
& Entity presence, size assessment, spatial relationship, and planning
& \checkmark 
& \checkmark \\
OmniSpatial~\citep{jia2025omnispatial} 
& Perspective-taking, spatial reasoning, interaction and logic
& 
& \checkmark \\
MindCube~\citep{yin2025spatial} 
& Perspective-taking, cognitive mapping, and mental simulation  
& 
& \checkmark \\
\bottomrule
\label{tab:benchmarks}
\end{tabular}
}
\end{table}

\paragraph{Benchmarks.}
We evaluate SpatialForge on five complementary spatial reasoning benchmarks: CV-Bench~\citep{tong2024cambrian}, SPAR-Bench~\citep{zhang2025flatland}, SpaCE10~\citep{Gong2025SpaCE10AC}, OmniSpatial~\citep{jia2025omnispatial}, and MindCube~\citep{yin2025spatial}, as summarized in Table~\ref{tab:benchmarks}. 
Together, these benchmarks cover a progression from object-level spatial perception and pairwise geometric relations to compositional reasoning and viewpoint-dependent spatial cognition. 
This evaluation suite allows us to test whether SpatialForge improves general spatial ability rather than overfitting to a single relation type or benchmark format.
Importantly, SpatialForge is not designed for any specific benchmark. Instead, it is constructed through a unified pipeline that decomposes spatial understanding into perception and relation, using open-world 2D images. As a result, our supervision naturally overlaps with the capabilities required by these benchmarks, enabling them to serve as unbiased evaluation tools for assessing generalizable spatial reasoning.

\begin{table}[t]
    \centering
    \caption{
\textbf{Results on spatial reasoning benchmarks} The benchmarks include CV-Bench~\citep{tong2024cambrian}, SPAR-Bench~\citep{zhang2025flatland}, SpaCE10~\citep{Gong2025SpaCE10AC}, OmniSpatial~\citep{jia2025omnispatial}, and MindCube~\citep{yin2025spatial}. \textbf{Bold} and \underline{underline} mark the best and second-best results among open-source baselines and our model.
Numbers in parentheses indicate the absolute improvement over the base model Qwen3-VL-2B.
}
    \label{tab:main_result}
    \resizebox{\textwidth}{!}{%
    \setlength{\tabcolsep}{0.6pt}
    \tiny
    \begin{tabular}{lcccccccc}
    \toprule
    \multirow{2}{*}{Methods} 
    & \multicolumn{3}{c}{CV-Bench~\citep{tong2024cambrian}}
    & \multirow{2}{*}{SPAR~\citep{zhang2025flatland}} 
    & \multirow{2}{*}{SpaCE10~\citep{Gong2025SpaCE10AC}} 
    & \multicolumn{2}{c}{OmniSpatial~\citep{jia2025omnispatial}} 
    & \multirow{2}{*}{MindCube~\citep{yin2025spatial}} \\
    
    \cmidrule(lr){2-4}
    \cmidrule(lr){7-8}
    
    & 2D & 3D & Avg. & & & Persp. & Avg. & \\
    
    \midrule
    Human Level & -- & -- & -- & 67.3 & 91.3 & 94.4 & 92.6 & -- \\
    Random & 25.0 & 25.0 & 25.0 & 32.7 & 25.0 & 33.6 & 24.9 & 32.4 \\
    
    \midrule
    \rowcolor{gray!10} Proprietary Models \\
    
    Gemini-2.0-Flash-Thinking~\citep{team2023gemini} & -- & -- & -- & -- & 34.3 & 47.4 & 44.0 & 47.1 \\
    GPT-4o~\citep{achiam2023gpt} &69.4 & 81.3 & 75.4 & 36.4 & 58.3 & 51.7 & 47.8 & -- \\
    Claude-3.7-Sonnet~\citep{anthropic2024claude3} & -- & -- & -- & 21.8 & 46.0 & 48.3 & 47.5 & -- \\
    
    \midrule
    \rowcolor{gray!10} Open-Source General Models \\
    LLaVA-OneVision-7B~\citep{li2024llava} & 53.2 & 63.5  & 58.3 & 31.2 & \textbf{45.2} & 40.2 &  35.7 & \textbf{47.4} \\
    InternVL3-2B~\citep{chen2024internvl} & -- & -- & -- & -- & \underline{44.2} & 41.2 & 38.0 & 37.5 \\
Qwen2.5-VL-3B~\citep{yang2025qwen3} & 69.1 & 72.2 &70.6 & 24.6 & 31.7 & 41.2 & \underline{40.3} & 33.2 \\
    Qwen2.5-VL-7B~\citep{yang2025qwen3} & \textbf{75.0} &83.1 & \textbf{79.0} & 39.2 & 37.4 & \underline{45.0} & 39.2 & 38.8 \\
    Qwen3-VL-2B~\citep{yang2025qwen3} & \underline{73.7} & \underline{83.4} & 78.6 & \underline{42.6} & 35.6 & 41.2 & 35.7 & 32.2 \\
    
    \midrule
    \rowcolor{gray!10} Open-sourced Specialized Models \\
    
    SpaceQwen2.5-VL-3B-Instruct\citep{chen2024spatialvlm} & 54.9 & 60.7 & 57.8 & 36.9 & 32.0 & \textbf{47.4} & \underline{40.3} & 33.3 \\
    Spatial-MLLM-4B~\citep{wu2025spatial} & -- & -- & --& 31.5 & -- & -- & -- & 32.1 \\
    SpaceR-7B~\citep{ouyang2025spacer} & 49.9 & 36.4 & 43.2 & 37.6 & 33.3 & 32.1 & 30.3 & 37.9 \\
    SpaceMantis-8B~\citep{jiang2024mantis} & -- & -- &  -- &  41.0 & 26.3 & 42.3 & 36.4 & 22.8 \\
    SpatialBot-3B~\citep{cai2025spatialbot} & -- & -- & -- & -- & --& 40.2 & 35.7 & -- \\
   SpatialLadder-3B~\citep{li2025spatialladder} & 72.4 &74.9 & 73.7 & 34.4 & 27.1 & 27.5 & 26.9 & 32.5 \\
    
    \midrule
    \rowcolor{gray!10} Ours \\
    
    \textbf{SpatialForge\textsubscript{Qwen3-VL-2B} } 
    & 72.2\textsubscript{(-1.5)} 
    & \textbf{85.2\textsubscript{(+1.8)}}
    & \underline{78.7\textsubscript{(+0.1)}}
    & \textbf{50.6\textsubscript{(+8.0)}} 
    & 38.0\textsubscript{(+2.4)} 
    & 44.2\textsubscript{(+3.0)} 
    & \textbf{43.1\textsubscript{(+7.4)} }
    & \underline{42.3\textsubscript{(+10.1)}} \\
    
    \bottomrule
    \end{tabular}%
    }
\end{table}

\subsection{Evaluation on Existing Spatial Benchmarks}
\label{evaluation_spatial}
\textbf{Spatial perception and relation reasoning.}
We evaluate SpatialForge on CV-Bench, SPAR-Bench, and SpaCE10, which cover  spatial perception, depth-aware relation reasoning, and compositional spatial understanding. 
As shown in Table~\ref{tab:main_result}, SpatialForge yields the notable improvements on relation-heavy and depth-sensitive benchmarks, improving CV-Bench-3D from 83.4 to 85.2 and SPAR-Bench from 42.6 to 50.6. 
This suggests that the model effectively learns geometric relationships such as depth ordering and relative spatial arrangement.
Notably, SpatialForge also improves performance on SpaCE10, which requires more compositional spatial reasoning beyond pairwise comparisons. Since our training data primarily provides pairwise spatial supervision, this gain indicates that the learned geometric primitives can transfer to more complex reasoning scenarios, rather than being limited to the training task format.
On CV-Bench-2D, we observe a slight performance drop, which we attribute to differences in annotation format. CV-Bench-2D relies on color-coded visual markers to indicate candidate regions, whereas SpatialForge is trained with natural language descriptions and box-based grounding. 
Overall, the results show that SpatialForge consistently improves spatial reasoning performance across diverse benchmarks.

\textbf{Viewpoint-dependent spatial reasoning.}
We further evaluate SpatialForge on OmniSpatial and MindCube, which emphasize viewpoint-dependent spatial reasoning and perspective-taking. 
SpatialForge improves the OmniSpatial average from 35.7 to 43.1 and MindCube from 32.2 to 42.3. 
These improvements suggest that the model can better handle changes in reference frames when reasoning about spatial relations. In particular, perspective-aware supervision helps the model correctly interpret relations such as \textit{left--right} under different viewpoints, rather than relying only on the default camera perspective. 
The gain on MindCube further indicates that the model can infer spatial relationships even when they are not directly aligned with the visible layout, requiring implicit reasoning over viewpoint changes.
\vspace{-2mm}

\subsection{Ablation Study}
\label{sec:ablation}
To investigate the contribution of each spatial data component, we perform controlled comparisons under a unified full-parameter fine-tuning setting based on Qwen3-VL-2B. 
As shown in Table~\ref{tab:ablation}, we compare four variants: spatial perception data only, relation-only data, perception combined with basic relation data, and the full SpatialForge setting. 
All variants are trained under identical settings to ensure fair comparison.

\paragraph{Single-component training.}
Fine-tuning on a single type of spatial data leads to degraded or unstable performance compared to the baseline, likely due to reduced data diversity and distribution shift. This suggests that neither perception nor relation signals alone is sufficient for robust spatial reasoning. 
Perception-only training is also sensitive to annotation/interface mismatch (e.g., CV-Bench-2D uses color-coded markers, while our supervision relies on language and boxes), which may further contribute to the drop.

\paragraph{Complementarity of perception and relation.}
Combining spatial perception with basic relation data leads to substantial improvements over the individual components, demonstrating the complementary roles of these signals. 
Perception data supports object localization and recognition, while relation data introduces geometric constraints such as \textit{near--far} and \textit{left--right} comparisons. 
Their combination enables the model to jointly capture object-level information and inter-object spatial relationships, resulting in consistently stronger performance across benchmarks.

\paragraph{Effect of perspective-aware supervision.}
Building upon this, incorporating \textit{perspective}-aware supervision further improves performance over the perception + basic relation setting. 
This indicates that perspective information provides additional cues beyond pairwise geometric relations. 
By introducing viewpoint-dependent reasoning, the model learns to interpret spatial relations under different reference frames rather than relying solely on image-centric coordinates. 
This is particularly beneficial for viewpoint-sensitive benchmarks such as OmniSpatial and MindCube, where correct reasoning depends on the adopted reference frame.

\begin{table}[t]
\centering
\caption{\textbf{Ablation study on spatial data components.}
\textit{Perc.} denotes spatial perception data; \textit{Basic Rel.} uses \textit{near--far} and \textit{left--right} relations; and \textit{Rel.} further incorporates \textit{perspective}-aware relations.}
\label{tab:ablation}
\resizebox{\textwidth}{!}{%
\setlength{\tabcolsep}{4pt}
\small
\begin{tabular}{lcccccccc}
\toprule
    \multirow{2}{*}{Methods} 
    & \multicolumn{3}{c}{CV-Bench~\citep{tong2024cambrian}}
    & \multirow{2}{*}{SPAR~\citep{zhang2025flatland}} 
    & \multirow{2}{*}{SpaCE10~\citep{Gong2025SpaCE10AC}} 
    & \multicolumn{2}{c}{OmniSpatial~\citep{jia2025omnispatial}} 
    & \multirow{2}{*}{MindCube~\citep{yin2025spatial}} \\
    
    \cmidrule(lr){2-4}
    \cmidrule(lr){7-8}
& 2D & 3D & Avg. & & & Persp. & Avg. & \\
\midrule
Baseline (Qwen3-VL-2B)
& \textbf{73.7} & \underline{83.4} & \underline{78.6} & 42.6 & \underline{35.6} & 41.2 & 35.7 & 32.2 \\

\midrule
\multicolumn{9}{l}{\textit{+ Spatial Data}} \\
\quad + Perc.
& 62.7 & 73.0 & 67.9 & 31.2 & 29.4 & 43.0 & 37.1 & 31.2 \\

\quad + Rel.
& 65.4 & 80.2 & 72.8 & \underline{46.2} & 31.6 & 41.2 & 39.5 & 33.9 \\

\quad + Perc. + Basic Rel.
& 70.5 & 83.0 & 76.8 & 45.9 & 35.3 & \underline{43.1} & \underline{41.6} & \underline{37.5} \\

\quad + Perc. + Rel.
& \underline{72.2} & \textbf{85.2} & \textbf{78.7} & \textbf{50.6} & \textbf{38.0} & \textbf{44.2} & \textbf{43.1} & \textbf{42.3} \\
\bottomrule
\end{tabular}%
}
\end{table}

\section{Conclusion and Limitations}

In this paper, we presented SpatialForge, a data-centric framework for enhancing the spatial reasoning capabilities of VLMs.
By shifting from hard-to-scale scene-centric annotations to a scalable 2D-driven data synthesis strategy, we bridge the gap between semantic understanding and geometric reasoning. 
Our SpatialForge-10M dataset demonstrates that large-scale and well-structured spatial supervision can serve as an effective signal for learning spatial awareness.
Through the integration of spatial perception and relation tasks, our framework enables VLMs to better capture depth, layout, and viewpoint-dependent relationships from diverse in-the-wild images. 
Experimental results show consistent improvements across multiple spatial reasoning benchmarks, indicating strong generalization to different spatial settings.
Overall, our findings suggest that scaling 2D spatial supervision is a practical and effective direction for improving 3D-aware understanding in VLMs. 

Our approach has several limitations. First, spatial supervision inferred from single-view images is inherently approximate, which may affect accuracy in complex scenarios. Second, the multi-stage data synthesis pipeline can introduce noise despite verification. Third, our current design focuses on a limited set of spatial relations and does not yet cover more complex reasoning such as physical interactions or temporal dynamics. Finally, as our method does not rely on explicit 3D data, it may be less suitable for tasks requiring precise metric geometry. We provide a more detailed discussion in Appendix~\ref{appendix:limitation}.
\bibliographystyle{unsrt}
\bibliography{reference}

@article{tong2024cambrian,
  title={Cambrian-1: A fully open, vision-centric exploration of multimodal llms},
  author={Tong, Shengbang and Brown, Ellis and Wu, Penghao and Woo, Sanghyun and Middepogu, Manoj and Akula, Sai C and Yang, Jihan and Yang, Shusheng and Iyer, Adithya and Pan, Xichen and others},
  journal={Advances in Neural Information Processing Systems},
  volume={37},
  pages={87310--87356},
  year={2024}
}

@inproceedings{tong2024eyes,
  title={Eyes wide shut? exploring the visual shortcomings of multimodal llms},
  author={Tong, Shengbang and Liu, Zhuang and Zhai, Yuexiang and Ma, Yi and LeCun, Yann and Xie, Saining},
  booktitle={Proceedings of the IEEE/CVF conference on computer vision and pattern recognition},
  pages={9568--9578},
  year={2024}
}

@misc{zhu2025llava3dsimpleeffectivepathway,
      title={LLaVA-3D: A Simple yet Effective Pathway to Empowering LMMs with 3D-awareness}, 
      author={Chenming Zhu and Tai Wang and Wenwei Zhang and Jiangmiao Pang and Xihui Liu},
      year={2025},
      eprint={2409.18125},
      archivePrefix={arXiv},
      primaryClass={cs.CV},
      url={https://arxiv.org/abs/2409.18125}, 
}

@article{liu2023visual,
  title={Visual instruction tuning},
  author={Liu, Haotian and Li, Chunyuan and Wu, Qingyang and Lee, Yong Jae},
  journal={Advances in neural information processing systems},
  volume={36},
  pages={34892--34916},
  year={2023}
}

@inproceedings{liu2024grounding,
  title={Grounding dino: Marrying dino with grounded pre-training for open-set object detection},
  author={Liu, Shilong and Zeng, Zhaoyang and Ren, Tianhe and Li, Feng and Zhang, Hao and Yang, Jie and Jiang, Qing and Li, Chunyuan and Yang, Jianwei and Su, Hang and others},
  booktitle={European conference on computer vision},
  pages={38--55},
  year={2024},
  organization={Springer}
}

@article{yang2024depth,
  title={Depth anything v2},
  author={Yang, Lihe and Kang, Bingyi and Huang, Zilong and Zhao, Zhen and Xu, Xiaogang and Feng, Jiashi and Zhao, Hengshuang},
  journal={Advances in Neural Information Processing Systems},
  volume={37},
  pages={21875--21911},
  year={2024}
}

@article{wang2024picture,
  title={Is a picture worth a thousand words? delving into spatial reasoning for vision language models},
  author={Wang, Jiayu and Ming, Yifei and Shi, Zhenmei and Vineet, Vibhav and Wang, Xin and Li, Yixuan and Joshi, Neel},
  journal={Advances in Neural Information Processing Systems},
  volume={37},
  pages={75392--75421},
  year={2024}
}

@inproceedings{chen2024spatialvlm,
  title={Spatialvlm: Endowing vision-language models with spatial reasoning capabilities},
  author={Chen, Boyuan and Xu, Zhuo and Kirmani, Sean and Ichter, Brain and Sadigh, Dorsa and Guibas, Leonidas and Xia, Fei},
  booktitle={Proceedings of the IEEE/CVF Conference on Computer Vision and Pattern Recognition},
  pages={14455--14465},
  year={2024}
}

@article{jiang2024mantis,
  title={MANTIS: Interleaved Multi-Image Instruction Tuning},
  author={Jiang, Dongfu and He, Xuan and Zeng, Huaye and Wei, Con and Ku, Max and Liu, Qian and Chen, Wenhu},
  journal={arXiv preprint arXiv:2405.01483},
  year={2024}
}

@inproceedings{liu2024mmbench,
  title={Mmbench: Is your multi-modal model an all-around player?},
  author={Liu, Yuan and Duan, Haodong and Zhang, Yuanhan and Li, Bo and Zhang, Songyang and Zhao, Wangbo and Yuan, Yike and Wang, Jiaqi and He, Conghui and Liu, Ziwei and others},
  booktitle={European conference on computer vision},
  pages={216--233},
  year={2024},
  organization={Springer}
}

@inproceedings{mathew2021docvqa,
  title={Docvqa: A dataset for vqa on document images},
  author={Mathew, Minesh and Karatzas, Dimosthenis and Jawahar, CV},
  booktitle={Proceedings of the IEEE/CVF winter conference on applications of computer vision},
  pages={2200--2209},
  year={2021}
}

@article{fu2024ocrbench,
  title={Ocrbench v2: An improved benchmark for evaluating large multimodal models on visual text localization and reasoning},
  author={Fu, Ling and Kuang, Zhebin and Song, Jiajun and Huang, Mingxin and Yang, Biao and Li, Yuzhe and Zhu, Linghao and Luo, Qidi and Wang, Xinyu and Lu, Hao and others},
  journal={arXiv preprint arXiv:2501.00321},
  year={2024}
}

@inproceedings{zitkovich2023rt,
  title={Rt-2: Vision-language-action models transfer web knowledge to robotic control},
  author={Zitkovich, Brianna and Yu, Tianhe and Xu, Sichun and Xu, Peng and Xiao, Ted and Xia, Fei and Wu, Jialin and Wohlhart, Paul and Welker, Stefan and Wahid, Ayzaan and others},
  booktitle={Conference on Robot Learning},
  pages={2165--2183},
  year={2023},
  organization={PMLR}
}

@article{tian2024drivevlm,
  title={Drivevlm: The convergence of autonomous driving and large vision-language models},
  author={Tian, Xiaoyu and Gu, Junru and Li, Bailin and Liu, Yicheng and Wang, Yang and Zhao, Zhiyong and Zhan, Kun and Jia, Peng and Lang, Xianpeng and Zhao, Hang},
  journal={arXiv preprint arXiv:2402.12289},
  year={2024}
}

@article{chandrasegaran2024hourvideo,
  title={Hourvideo: 1-hour video-language understanding},
  author={Chandrasegaran, Keshigeyan and Gupta, Agrim and Hadzic, Lea M and Kota, Taran and He, Jimming and Eyzaguirre, Crist{\'o}bal and Durante, Zane and Li, Manling and Wu, Jiajun and Fei-Fei, Li},
  journal={Advances in Neural Information Processing Systems},
  volume={37},
  pages={53168--53197},
  year={2024}
}

@article{deng2025internspatial,
  title={Internspatial: A comprehensive dataset for spatial reasoning in vision-language models},
  author={Deng, Nianchen and Gu, Lixin and Ye, Shenglong and He, Yinan and Chen, Zhe and Li, Songze and Wang, Haomin and Wei, Xingguang and Yang, Tianshuo and Dou, Min and others},
  journal={arXiv preprint arXiv:2506.18385},
  year={2025}
}

@inproceedings{song2025robospatial,
  title={Robospatial: Teaching spatial understanding to 2d and 3d vision-language models for robotics},
  author={Song, Chan Hee and Blukis, Valts and Tremblay, Jonathan and Tyree, Stephen and Su, Yu and Birchfield, Stan},
  booktitle={Proceedings of the Computer Vision and Pattern Recognition Conference},
  pages={15768--15780},
  year={2025}
}

@article{tian2026last,
  title={LAST: Leveraging Tools as Hints to Enhance Spatial Reasoning for Multimodal Large Language Models},
  author={Tian, Shi-Yu and Zhou, Zhi and Yu, Kun-Yang and Yang, Ming and Chen, Yang and Shang, Ziqiao and Guo, Lan-Zhe and Li, Yu-Feng},
  journal={arXiv preprint arXiv:2604.09712},
  year={2026}
}

@inproceedings{suris2023vipergpt,
  title={Vipergpt: Visual inference via python execution for reasoning},
  author={Sur{\'\i}s, D{\'\i}dac and Menon, Sachit and Vondrick, Carl},
  booktitle={Proceedings of the IEEE/CVF international conference on computer vision},
  pages={11888--11898},
  year={2023}
}

@article{driess2023palm,
  title={Palm-e: An embodied multimodal language model},
  author={Driess, Danny and Xia, Fei and Sajjadi, Mehdi SM and Lynch, Corey and Chowdhery, Aakanksha and Ichter, Brian and Wahid, Ayzaan and Tompson, Jonathan and Vuong, Quan and Yu, Tianhe and others},
  journal={arXiv preprint arXiv:2303.03378},
  year={2023}
}

@inproceedings{yue2024mmmu,
  title={Mmmu: A massive multi-discipline multimodal understanding and reasoning benchmark for expert agi},
  author={Yue, Xiang and Ni, Yuansheng and Zhang, Kai and Zheng, Tianyu and Liu, Ruoqi and Zhang, Ge and Stevens, Samuel and Jiang, Dongfu and Ren, Weiming and Sun, Yuxuan and others},
  booktitle={Proceedings of the IEEE/CVF conference on computer vision and pattern recognition},
  pages={9556--9567},
  year={2024}
}

@article{li2024llava,
  title={Llava-onevision: Easy visual task transfer},
  author={Li, Bo and Zhang, Yuanhan and Guo, Dong and Zhang, Renrui and Li, Feng and Zhang, Hao and Zhang, Kaichen and Zhang, Peiyuan and Li, Yanwei and Liu, Ziwei and others},
  journal={arXiv preprint arXiv:2408.03326},
  year={2024}
}

@article{achiam2023gpt,
  title={Gpt-4 technical report},
  author={Achiam, Josh and Adler, Steven and Agarwal, Sandhini and Ahmad, Lama and Akkaya, Ilge and Aleman, Florencia Leoni and Almeida, Diogo and Altenschmidt, Janko and Altman, Sam and Anadkat, Shyamal and others},
  journal={arXiv preprint arXiv:2303.08774},
  year={2023}
}

@article{zhang2024ferret,
  title={Ferret-v2: An improved baseline for referring and grounding with large language models},
  author={Zhang, Haotian and You, Haoxuan and Dufter, Philipp and Zhang, Bowen and Chen, Chen and Chen, Hong-You and Fu, Tsu-Jui and Wang, William Yang and Chang, Shih-Fu and Gan, Zhe and others},
  journal={arXiv preprint arXiv:2404.07973},
  year={2024}
}

@inproceedings{ranasinghe2024learning,
  title={Learning to localize objects improves spatial reasoning in visual-llms},
  author={Ranasinghe, Kanchana and Shukla, Satya Narayan and Poursaeed, Omid and Ryoo, Michael S and Lin, Tsung-Yu},
  booktitle={Proceedings of the IEEE/CVF Conference on Computer Vision and Pattern Recognition},
  pages={12977--12987},
  year={2024}
}

@article{chen2023shikra,
  title={Shikra: Unleashing multimodal llm's referential dialogue magic},
  author={Chen, Keqin and Zhang, Zhao and Zeng, Weili and Zhang, Richong and Zhu, Feng and Zhao, Rui},
  journal={arXiv preprint arXiv:2306.15195},
  year={2023}
}

@article{hong20233d,
  title={3d-llm: Injecting the 3d world into large language models},
  author={Hong, Yining and Zhen, Haoyu and Chen, Peihao and Zheng, Shuhong and Du, Yilun and Chen, Zhenfang and Gan, Chuang},
  journal={Advances in Neural Information Processing Systems},
  volume={36},
  pages={20482--20494},
  year={2023}
}

@inproceedings{xu2024pointllm,
  title={Pointllm: Empowering large language models to understand point clouds},
  author={Xu, Runsen and Wang, Xiaolong and Wang, Tai and Chen, Yilun and Pang, Jiangmiao and Lin, Dahua},
  booktitle={European Conference on Computer Vision},
  pages={131--147},
  year={2024},
  organization={Springer}
}

@inproceedings{qi2024shapellm,
  title={Shapellm: Universal 3d object understanding for embodied interaction},
  author={Qi, Zekun and Dong, Runpei and Zhang, Shaochen and Geng, Haoran and Han, Chunrui and Ge, Zheng and Yi, Li and Ma, Kaisheng},
  booktitle={European Conference on Computer Vision},
  pages={214--238},
  year={2024},
  organization={Springer}
}

@inproceedings{yang2025thinking,
  title={Thinking in space: How multimodal large language models see, remember, and recall spaces},
  author={Yang, Jihan and Yang, Shusheng and Gupta, Anjali W and Han, Rilyn and Fei-Fei, Li and Xie, Saining},
  booktitle={Proceedings of the Computer Vision and Pattern Recognition Conference},
  pages={10632--10643},
  year={2025}
}

@article{jia2025omnispatial,
  title={Omnispatial: Towards comprehensive spatial reasoning benchmark for vision language models},
  author={Jia, Mengdi and Qi, Zekun and Zhang, Shaochen and Zhang, Wenyao and Yu, Xinqiang and He, Jiawei and Wang, He and Yi, Li},
  journal={arXiv preprint arXiv:2506.03135},
  year={2025}
}

@article{zhang2025flatland,
  title={From flatland to space: Teaching vision-language models to perceive and reason in 3d},
  author={Zhang, Jiahui and Chen, Yurui and Zhou, Yanpeng and Xu, Yueming and Huang, Ze and Mei, Jilin and Chen, Junhui and Yuan, Yu-Jie and Cai, Xinyue and Huang, Guowei and others},
  journal={arXiv preprint arXiv:2503.22976},
  year={2025}
}

@article{cheng2024spatialrgpt,
  title={Spatialrgpt: Grounded spatial reasoning in vision-language models},
  author={Cheng, An-Chieh and Yin, Hongxu and Fu, Yang and Guo, Qiushan and Yang, Ruihan and Kautz, Jan and Wang, Xiaolong and Liu, Sifei},
  journal={Advances in Neural Information Processing Systems},
  volume={37},
  pages={135062--135093},
  year={2024}
}

@article{li2024proximity,
  title={Proximity qa: Unleashing the power of multi-modal large language models for spatial proximity analysis},
  author={Li, Jianing and Nan, Xi and Lu, Ming and Du, Li and Zhang, Shanghang},
  journal={arXiv preprint arXiv:2401.17862},
  year={2024}
}

@inproceedings{cai2025spatialbot,
  title={Spatialbot: Precise spatial understanding with vision language models},
  author={Cai, Wenxiao and Ponomarenko, Iaroslav and Yuan, Jianhao and Li, Xiaoqi and Yang, Wankou and Dong, Hao and Zhao, Bo},
  booktitle={2025 IEEE International Conference on Robotics and Automation (ICRA)},
  pages={9490--9498},
  year={2025},
  organization={IEEE}
}

@article{ouyang2025spacer,
  title={Spacer: Reinforcing mllms in video spatial reasoning},
  author={Ouyang, Kun and Liu, Yuanxin and Wu, Haoning and Liu, Yi and Zhou, Hao and Zhou, Jie and Meng, Fandong and Sun, Xu},
  journal={arXiv preprint arXiv:2504.01805},
  year={2025}
}

@article{li2025spatialladder,
  title={Spatialladder: Progressive training for spatial reasoning in vision-language models},
  author={Li, Hongxing and Li, Dingming and Wang, Zixuan and Yan, Yuchen and Wu, Hang and Zhang, Wenqi and Shen, Yongliang and Lu, Weiming and Xiao, Jun and Zhuang, Yueting},
  journal={arXiv preprint arXiv:2510.08531},
  year={2025}
}

@inproceedings{dai2017scannet,
  title={Scannet: Richly-annotated 3d reconstructions of indoor scenes},
  author={Dai, Angela and Chang, Angel X and Savva, Manolis and Halber, Maciej and Funkhouser, Thomas and Nie{\ss}ner, Matthias},
  booktitle={Proceedings of the IEEE conference on computer vision and pattern recognition},
  pages={5828--5839},
  year={2017}
}

@inproceedings{yeshwanth2023scannet++,
  title={Scannet++: A high-fidelity dataset of 3d indoor scenes},
  author={Yeshwanth, Chandan and Liu, Yueh-Cheng and Nie{\ss}ner, Matthias and Dai, Angela},
  booktitle={Proceedings of the IEEE/CVF International Conference on Computer Vision},
  pages={12--22},
  year={2023}
}

@article{wu2025spatial,
  title={Spatial-mllm: Boosting mllm capabilities in visual-based spatial intelligence},
  author={Wu, Diankun and Liu, Fangfu and Hung, Yi-Hsin and Duan, Yueqi},
  journal={arXiv preprint arXiv:2505.23747},
  year={2025}
}

@inproceedings{yin2025spatial,
  title={Spatial mental modeling from limited views},
  author={Yin, Baiqiao and Wang, Qineng and Zhang, Pingyue and Zhang, Jianshu and Wang, Kangrui and Wang, Zihan and Zhang, Jieyu and Chandrasegaran, Keshigeyan and Liu, Han and Krishna, Ranjay and others},
  booktitle={Structural Priors for Vision Workshop at ICCV'25},
  year={2025}
}

@article{Gong2025SpaCE10AC,
  title={SpaCE-10: A Comprehensive Benchmark for Multimodal Large Language Models in Compositional Spatial Intelligence},
  author={Ziyang Gong and Wenhao Li and Olivera Mart{\'i}nez Ma and Songyuan Li and Jiayi Ji and Xue Yang and Gen Luo and Junchi Yan and Rongrong Ji},
  journal={ArXiv},
  year={2025},
  volume={abs/2506.07966},
  url={https://api.semanticscholar.org/CorpusID:279251735}
}

@article{li2025imagine,
  title={Imagine while reasoning in space: Multimodal visualization-of-thought},
  author={Li, Chengzu and Wu, Wenshan and Zhang, Huanyu and Xia, Yan and Mao, Shaoguang and Dong, Li and Vuli{\'c}, Ivan and Wei, Furu},
  journal={arXiv preprint arXiv:2501.07542},
  year={2025}
}

@article{liu2025spatialcot,
  title={Spatialcot: Advancing spatial reasoning through coordinate alignment and chain-of-thought for embodied task planning},
  author={Liu, Yuecheng and Chi, Dafeng and Wu, Shiguang and Zhang, Zhanguang and Hu, Yaochen and Zhang, Lingfeng and Zhang, Yingxue and Wu, Shuang and Cao, Tongtong and Huang, Guowei and others},
  journal={arXiv preprint arXiv:2501.10074},
  year={2025}
}

@inproceedings{shao2019objects365,
  title={Objects365: A large-scale, high-quality dataset for object detection},
  author={Shao, Shuai and Li, Zeming and Zhang, Tianyuan and Peng, Chao and Yu, Gang and Zhang, Xiangyu and Li, Jing and Sun, Jian},
  booktitle={Proceedings of the IEEE/CVF international conference on computer vision},
  pages={8430--8439},
  year={2019}
}

@inproceedings{deitke2025molmo,
  title={Molmo and pixmo: Open weights and open data for state-of-the-art vision-language models},
  author={Deitke, Matt and Clark, Christopher and Lee, Sangho and Tripathi, Rohun and Yang, Yue and Park, Jae Sung and Salehi, Mohammadreza and Muennighoff, Niklas and Lo, Kyle and Soldaini, Luca and others},
  booktitle={Proceedings of the Computer Vision and Pattern Recognition Conference},
  pages={91--104},
  year={2025}
}

@article{kuznetsova2020open,
  title={The open images dataset v4: Unified image classification, object detection, and visual relationship detection at scale},
  author={Kuznetsova, Alina and Rom, Hassan and Alldrin, Neil and Uijlings, Jasper and Krasin, Ivan and Pont-Tuset, Jordi and Kamali, Shahab and Popov, Stefan and Malloci, Matteo and Kolesnikov, Alexander and others},
  journal={International journal of computer vision},
  volume={128},
  number={7},
  pages={1956--1981},
  year={2020},
  publisher={Springer}
}

@article{yang2025qwen3,
  title={Qwen3 technical report},
  author={Yang, An and Li, Anfeng and Yang, Baosong and Zhang, Beichen and Hui, Binyuan and Zheng, Bo and Yu, Bowen and Gao, Chang and Huang, Chengen and Lv, Chenxu and others},
  journal={arXiv preprint arXiv:2505.09388},
  year={2025}
}

@inproceedings{chen2024internvl,
  title={Internvl: Scaling up vision foundation models and aligning for generic visual-linguistic tasks},
  author={Chen, Zhe and Wu, Jiannan and Wang, Wenhai and Su, Weijie and Chen, Guo and Xing, Sen and Zhong, Muyan and Zhang, Qinglong and Zhu, Xizhou and Lu, Lewei and others},
  booktitle={Proceedings of the IEEE/CVF conference on computer vision and pattern recognition},
  pages={24185--24198},
  year={2024}
}

@article{team2023gemini,
  title={Gemini: a family of highly capable multimodal models},
  author={Team, Gemini and Anil, Rohan and Borgeaud, Sebastian and Alayrac, Jean-Baptiste and Yu, Jiahui and Soricut, Radu and Schalkwyk, Johan and Dai, Andrew M and Hauth, Anja and Millican, Katie and others},
  journal={arXiv preprint arXiv:2312.11805},
  year={2023}
}

@techreport{anthropic2024claude3,
  title       = {The Claude 3 Model Family: Opus, Sonnet, Haiku},
  author      = {{Anthropic}},
  year        = {2024},
  institution = {Anthropic},
  type        = {Model Card},
  url         = {https://www-cdn.anthropic.com/de8ba9b01c9ab7cbabf5c33b80b7bbc618857627/Model_Card_Claude_3.pdf}
}

@inproceedings{radford2021learning,
  title={Learning transferable visual models from natural language supervision},
  author={Radford, Alec and Kim, Jong Wook and Hallacy, Chris and Ramesh, Aditya and Goh, Gabriel and Agarwal, Sandhini and Sastry, Girish and Askell, Amanda and Mishkin, Pamela and Clark, Jack and others},
  booktitle={International conference on machine learning},
  pages={8748--8763},
  year={2021},
  organization={PmLR}
}

\newpage

\section{Task Taxonomy}
\label{sec:appendix_taxonomy}

In this section, we provide a detailed breakdown of the spatial task taxonomy used in our dataset. 
We organize spatial reasoning into multiple capability levels and define each task with clear input-output formats.

\subsection{Task Definitions}
We present example task categories and templates in Table~\ref{tab:task_templates}. In practice, we leverage an LLM~\citep{achiam2023gpt} to generate a diverse set of templates; the table lists only a representative example for each category.
\begin{table}[h]
\centering
\caption{\textbf{Task categories and example templates in SpatialForge-10M.}}
\label{tab:task_templates}
\resizebox{\linewidth}{!}{
\small
\setlength{\tabcolsep}{3pt}
\renewcommand{\arraystretch}{1.2}
\begin{tabular}{p{2.2cm}p{5.5cm}p{5cm}}
\toprule
Task Category & Description & Example Template \\
\midrule
Grounding & Localize objects based on descriptions, output bounding boxes. & ``Describe the object <box>...</box> in details.'' \\
Referring & Find and localize objects described by natural language. & ``Help me find the \textbf{\textit{\{region\_caption\}}} / \textbf{\textit{\{category\}}}.'' \\
Counting & Count the number of objects or instances in the image. & ``How many \textbf{\textit{\{category\}}} can be seen in this photo?'' \\
Depth Reasoning & Order or compare objects based on distance from camera. & ``Order these objects from farthest to nearest from the viewer. Objects are A. \textbf{\textit{\{region\_caption\}}}, B. \textbf{\textit{\{region\_caption\}}}...'' \\
Left-Right & Determine left/right relationships from camera or person perspective. & ``From the camera viewpoint, who is positioned at the far right? Provide the bbox.'' \\
Perspective Taking & Reason about spatial relations from a specific human viewpoint. & ``Imagine you are standing in the same position and facing the same direction as \textbf{\textit{\{region\_caption\}}} located at <box>...</box>. Is \textbf{\textit{\{region\_caption\}}} located at <box>...</box> on this person's left or right?'' \\
\bottomrule
\end{tabular}
}
\end{table}

\begin{figure}[h]
  \centering
  \includegraphics[width=\columnwidth]{ 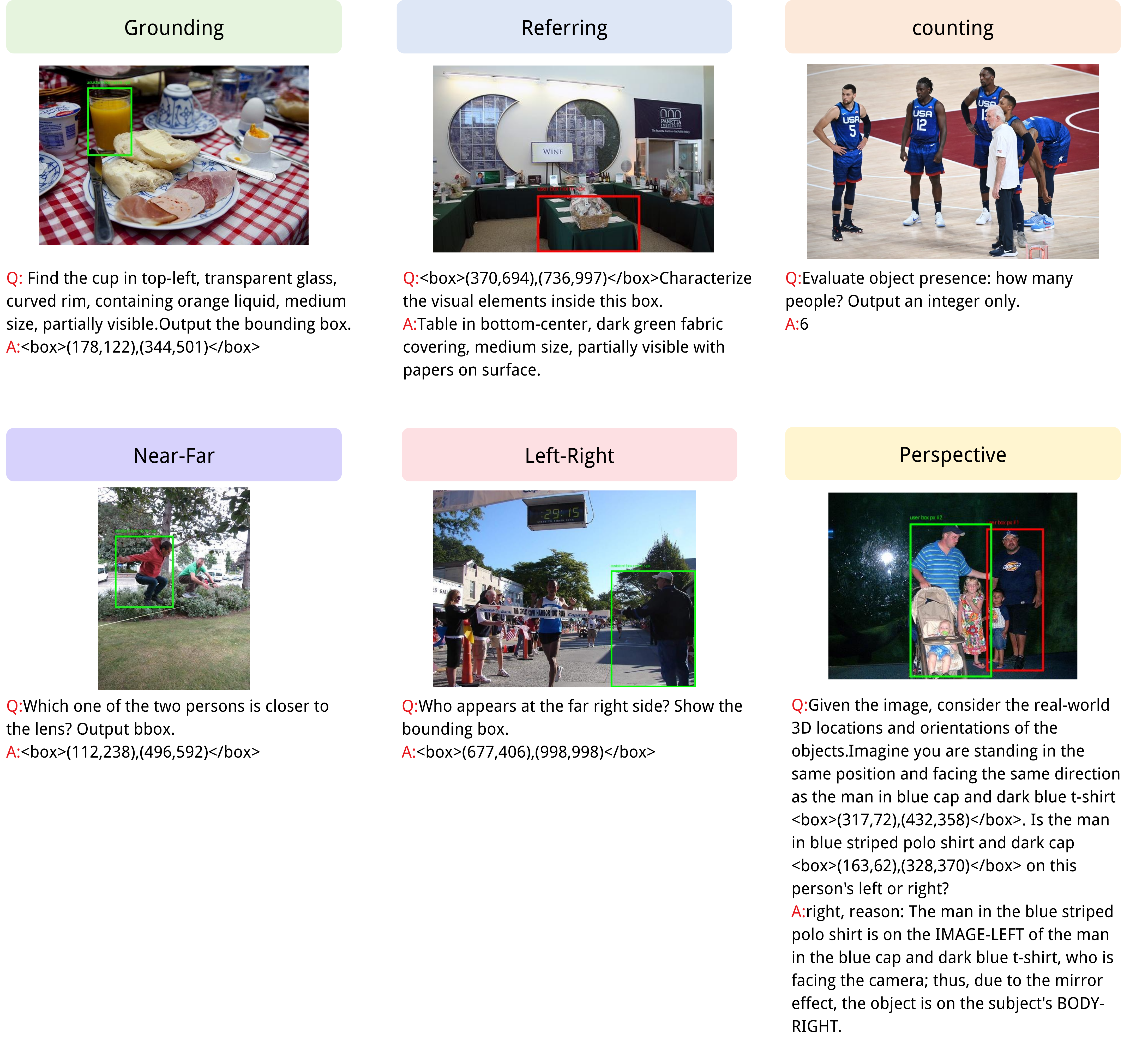}
  \caption{\textbf{Representative examples from six tasks in SpatialForge-10M.}}
    \label{fig:dataset_example}
\end{figure}

\begin{figure}[h]
  \centering
  \includegraphics[width=\columnwidth]{ 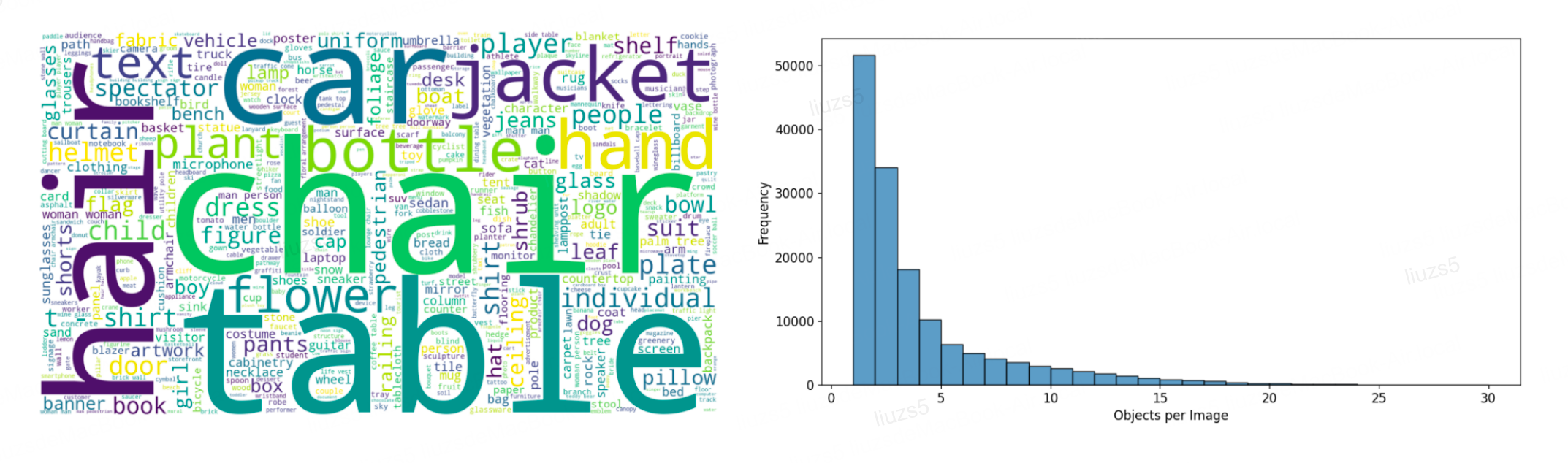}
  \caption{\textbf{Category Statistics of SpatialForge-10M dataset.} We present a word cloud visualization (left) and the distribution of object counts per image (right). The dataset exhibits broad coverage with high diversity and relatively balanced category distribution.}
    \label{fig:category_stats}
\end{figure}

\subsection{Statistics of SpatialForge-10M}
\label{appendix:statistics for dataset}
Table~\ref{tab:data_source_stats} provides a detailed breakdown of the dataset, showing the number of QA pairs for each task category and each data source. 
Figure~\ref{fig:category_stats} includes a word cloud of object categories and a bar chart showing the frequency distribution of object counts per image, demonstrating the semantic diversity of the dataset.
Additionally, Figure~\ref{fig:dataset_example} presents representative examples for each task, illustrating both the images and corresponding QA pairs. 

\begin{table}[h]
\centering
\caption{\textbf{Data statistics of SpatialForge-10M across different task categories and source data.}}
\label{tab:data_source_stats}
\resizebox{0.9\linewidth}{!}{
\small
\setlength{\tabcolsep}{3pt}
\renewcommand{\arraystretch}{1.1}
\begin{tabular}{lcccccccc}
\toprule
Source & Grounding & Referring & Counting & Near-Far & Left-Right & Persp. & Total \\
\midrule
Objects365   & 1,685,801 & 1,685,801 & -- & 2,527,378 & 20,970 & 3,981 & 5,923,931 \\
Pixmo        & 251,033   & 251,033   & -- & 32,379    & 72,443 & 454   & 607,342 \\
OpenImages   & 1,680,000 & 1,680,000 & 495,851 & -- & -- & 3,750 & 3,859,601 \\
\midrule
Total        & 3,616,834 & 3,616,834 & 495,851 & 2,559,757 & 93,413 & 8,185 & 10,190,874 \\
\bottomrule
\end{tabular}
}
\end{table}

\section{Details For Data Synthesis Pipeline}

\subsection{Image Filtering}
To ensure the quality and diversity of the training data, we apply a two-stage image filtering pipeline consisting of low-level quality filtering (e.g., resolution, exposure, sharpness) and high-level semantic selection using CLIP~\citep{radford2021learning} to retain only indoor and outdoor scenes while filtering out non-natural images such as documents, GUI screenshots, and asset renderings. The filtering statistics are summarized in Table~\ref{tab:filtering_stats}.

\begin{wrapfigure}{r}{0.45\textwidth}
\centering
\caption{\textbf{Image filtering statistics.}}
\label{tab:filtering_stats}
\small
\begin{tabular}{lcc}
\toprule
Data Source & Raw & Filtered \\
\midrule
Objects365 & 774,169 & 760,463 \\ 
OpenImages & 1,743,042 & 1,533,302 \\
Pixmo & 968,357 & 527,474 \\
\midrule
Total & 3,485,568 & 2,821,239 \\
\bottomrule
\end{tabular}
\end{wrapfigure}

\subsection{Data Preprocessing}
\label{appendix:data_preprocessing}

We construct object-centric representations through a hierarchical captioning and grounding pipeline. Given an input image, we employ Qwen3-VL-32B~\citep{yang2025qwen3} to generate both a detailed global caption that captures the overall scene semantics and a set of object categories (noun phrases) extracted from the caption. These categories serve as semantic queries for subsequent grounding. The queries are fed into an open-vocabulary detector (Grounding DINO~\citep{liu2024grounding}) to localize corresponding regions in the image, enabling flexible grounding beyond a fixed category set. Based on the resulting bounding boxes and associated object queries, we crop the corresponding image regions and feed them into Qwen3-VL-32B~\citep{yang2025qwen3} to generate fine-grained region-level captions. By focusing on the local visual content, this cropping strategy enables the model to produce more accurate and detailed descriptions. These region descriptions include detailed appearance, functional attributes, and spatial cues.

However, the open-vocabulary detector also introduces severe category imbalance. To address this, we apply a simple yet effective filtering strategy: (i) for overly frequent and semantically uninformative categories (e.g., \emph{sky}, \emph{tree}, \emph{window}, \emph{table}, \emph{floor}), we downsample them to 10\% of their original frequency to maintain diversity while reducing bias; (ii) for bounding boxes, we filter out those with aspect ratio outside $[1/3, 3]$ and those with area smaller than $100^2$ pixels, as such boxes are often detection noise or correspond to non-informative regions.

For depth estimation, we apply DepthAnythingV2~\citep{yang2024depth} to generate monocular depth maps for each image, providing dense per-pixel depth cues that facilitate understanding of occlusion, relative distance, and scene layout.
Furthermore, to support human-centric spatial reasoning (e.g., left/right from a person's perspective), we perform orientation estimation for human instances. Specifically, we use Qwen3-VL-32B~\citep{yang2025qwen3} to predict the facing direction of each detected person, restricted to a binary classification: facing toward the camera or facing away from the camera. This simplification is adopted because finer-grained orientation estimation (e.g., facing left, facing right, facing sideways) would involve more complex spatial reasoning that is difficult to reliably compute from in-the-wild 2D images. This binary orientation information is crucial for constructing perspective-aware QA pairs that require reasoning from a specific human viewpoint, enabling the model to perform allocentric left-right reasoning through the mirroring transformation.
The prompt used for VLMs are presented in Appendix~\ref{appendix:prompt for data_preprocess}.

\subsection{Task Workflow}
\label{appendix:task_workflow}

After data preprocessing, we construct QA pairs for six core task families.
\paragraph{Grounding and Referring.}
We generate two complementary types of QA pairs from the region-level annotations:
\begin{itemize}
    \item \textbf{Grounding (Bbox2Caption):} Given a bounding box, the model is asked to describe the object within it. The input format is ``Describe the object <box>...</box> in details.''
    \item \textbf{Referring (Caption2Bbox):} Given a region caption or category, the model is asked to localize the corresponding object. The input format is ``Help me find the \{\textit{region\_caption}\} / \{\textit{category}\}.''
\end{itemize}
Since we adopt Qwen3-VL~\citep{yang2025qwen3} as our base model and perform full-parameter fine-tuning, we normalize all bounding box coordinates to the range $[0, 1000]$ to align with its pre-training format. Specifically, for an image of width $W$ and height $H$, a bounding box with pixel coordinates $(x_{\min}, y_{\min}, x_{\max}, y_{\max})$ is normalized as:
\[
x' = \frac{x}{W} \times 1000, \quad y' = \frac{y}{H} \times 1000
\] 

\paragraph{Counting.}
Based on the extracted object categories, we generate counting QA pairs. To ensure meaningful supervision, we only retain questions where the object count in the image is greater than 1.

\paragraph{Near–Far.}
For each detected object region $R$ with bounding box $b$, we compute two complementary statistics from the predicted depth map $D$:
\begin{itemize}
    \item Median depth: $s_{\text{med}} = \underset{(x,y) \in R}{\text{median}} \; D(x,y)$, providing a stable estimate of the object's central depth.
    \item 90th percentile depth: $s_{p90} = \underset{(x,y) \in R}{P_{90}} \; D(x,y)$, capturing the far-side structure of the object.
\end{itemize}

Given two objects $A$ and $B$, we perform pairwise depth comparison using these two metrics. The relative depth ordering is determined as:
\[
\text{Order}(A,B) = 
\begin{cases}
A \prec B \text{ (A is nearer)} & \text{if } s_{\text{med}}(A) < s_{\text{med}}(B) \text{ and } s_{p90}(A) < s_{p90}(B) \\
B \prec A \text{ (B is nearer)} & \text{if } s_{\text{med}}(A) > s_{\text{med}}(B) \text{ and } s_{p90}(A) > s_{p90}(B) \\
\text{Ambiguous} & \text{otherwise}
\end{cases}
\]
where $A \prec B$ denotes that $A$ is closer to the camera than $B$.

Based on the agreement between the two metrics, we categorize each pair into four quality classes:
\begin{itemize}
    \item \textbf{Class A:} Both metrics are reliable and yield consistent ordering.
    \item \textbf{Class B:} Only $s_{\text{med}}$ is reliable (e.g., objects with high depth variance).
    \item \textbf{Class C:} Only $s_{p90}$ is reliable.
    \item \textbf{Class D:} Both metrics are reliable but yield inconsistent ordering.
\end{itemize}
Only pairs in Classes A–C are retained for depth reasoning QA generation.

\paragraph{Left-Right.}
We construct horizontal spatial relationships based on bounding box geometry. Given two objects $A$ and $B$ with bounding boxes $b_A = [x_A^{\min}, y_A^{\min}, x_A^{\max}, y_A^{\max}]$ and $b_B = [x_B^{\min}, y_B^{\min}, x_B^{\max}, y_B^{\max}]$, we propose a dual-anchor reasoning strategy to determine egocentric left–right ordering:

\begin{itemize}
    \item Center anchor: Compare the horizontal centers: $c_A = \frac{x_A^{\min} + x_A^{\max}}{2}$, $c_B = \frac{x_B^{\min} + x_B^{\max}}{2}$.
    \item Boundary anchor: Compare the left/right boundaries: $x_A^{\max}$ vs $x_B^{\min}$ (if $A$ is left of $B$) or $x_B^{\max}$ vs $x_A^{\min}$ (if $B$ is left of $A$).
\end{itemize}

The final ordering is determined as:
\[
\text{LeftRight}(A,B) = 
\begin{cases}
\text{Left} & \text{if } c_A < c_B \text{ and } x_A^{\max} < x_B^{\min} \\
\text{Right} & \text{if } c_A > c_B \text{ and } x_B^{\max} < x_A^{\min} \\
\text{Ambiguous} & \text{otherwise}
\end{cases}
\]
This dual-anchor design improves robustness in cases of partial overlap or varying object scales.

\paragraph{Perspective.}
Based on the estimated orientation, we derive the allocentric spatial relation $\mathcal{R}_{\text{allo}}$ as a function of the subject's viewpoint:
\begin{equation}
\mathcal{R}_{\text{allo}} = 
\begin{cases} 
    \mathcal{R}_{\text{ego}}, & \text{if } \theta_h = \text{away}, \\[1ex]
    \operatorname{reverse}(\mathcal{R}_{\text{ego}}), & \text{if } \theta_h = \text{toward},
\end{cases}
\end{equation}
where $\mathcal{R}_{\text{ego}}$ denotes the egocentric left-right relation (e.g., ``left'' or ``right'') determined by the dual-anchor strategy, and $\operatorname{reverse}(\cdot)$ flips the relation (i.e., $\operatorname{reverse}(\text{left}) = \text{right}$ and $\operatorname{reverse}(\text{right}) = \text{left}$).


This mirroring transformation enables the model to adopt the person's own egocentric perspective: when the person faces toward the camera, the model flips the left-right relations; when facing away, relations are preserved.

\subsection{Quality Inspector}
\label{appendix:quality}
To ensure the quality of our synthesized data, we employ a stronger VLM, Qwen3-VL-235B-A3B~\citep{yang2025qwen3}, as a quality inspector (prompt details are provided in the Appendix~\ref{appendix:prompt for data_preprocess}).For tasks where the answer is a bounding box (e.g., grounding and referring), we compute the Intersection over Union (IoU) between the predicted box and the ground-truth box, retaining only samples with IoU $\geq 0.8$. For tasks where the answer is a text string (e.g., depth ordering, left/right relations, and perspective taking), we apply exact string matching to verify consistency against the ground truth. Samples that fail the inspection are discarded. 



\section{Training Details}
\label{appendix:appendix_training}
We conduct supervised fine-tuning on the base model using full-parameter optimization, where all parameters—including the vision transformer (ViT), merger, and LLM—are updated. The training data comprises 10 million spatial QA pairs from our SpatialForge-10M dataset, augmented with 1 million general instruction-following samples from LLaVA-OneVision-1.5-Instruct-Data~\citep{li2024llava}. The model is trained on 32 NVIDIA H200 GPUs with a micro-batch size of 32, totaling approximately 24 hours of training time. Key hyperparameters are summarized in Table~\ref{tab:training_setting}.

\begin{table}[h]
\centering
\caption{\textbf{Training settings and hyperparameters for SpatialForge-2B.}}
\label{tab:training_setting}
\begin{tabular}{ll}
\toprule
Configurations & Values \\
\midrule
Base Model & Qwen3-VL-2B-Instruct \\
Learning Rate & 1.00e-05 \\
LR Decay Style & cosine \\
Epochs & 1 \\
Micro Batch Size & 32 \\
Max Seq Len & 4096 \\
Image Max Pixels & 602112 \\
Video Max Pixels & 602112 \\
Freeze ViT & false \\
Freeze Merger & false \\
Freeze LLM & false \\
\bottomrule
\end{tabular}
\end{table}

\section{Detailed Evaluation on Benchmarks}
\label{sec:appendix_results}
We conduct a detailed analysis on SPAR-Bench~\citep{zhang2025flatland} and MindCube~\citep{yin2025spatial}. For SPAR-Bench~\citep{zhang2025flatland} , we compare performance across its three task difficulty levels (Low, Medium, High) against baseline models, enabling a fine-grained evaluation of SpatialForge-2B's capabilities in spatial reasoning. For MindCube~\citep{yin2025spatial}, which strongly evaluates perspective transformation and mental rotation, we visualize representative cases to assess the model's ability in viewpoint switching and allocentric reasoning.

\subsection{Detailed Analysis on SPAR-Bench}
\label{appendix:spar}
Table~\ref{tab:detailed_results} provides a fine-grained breakdown of performance across three task levels in SPAR-Bench~\citep{zhang2025flatland}. Our SpatialForge-2B achieves 65.8 on low-level tasks, outperforming Qwen3-VL-2B (60.6), with strong object-object depth reasoning (Dist-OO: 73.1; Depth-OO: 68.2) and effective object-camera distance ordering (Dist-OC: 67.7 vs. 66.3). These improvements directly align with our depth-oriented QA pairs, which provide explicit numeric supervision (e.g., "order objects from farthest to nearest", "which object is closer to the camera?"). Although our dataset does not include absolute distance values, the relative distance ordering tasks offer effective numeric comparison signals. On medium-level tasks, we achieve 32.3, surpassing Qwen3-VL-2B (27.4). On high-level tasks, our model reaches 43.6, outperforming all baselines, with notable gains on DistI-OO (60.9 vs. 57.1), ObjRel-OO (53.6 vs. 45.6), and spatial imagination tasks. The performance on mental rotation and perspective-taking demonstrates that spatial supervision effectively enhances viewpoint transformation and allocentric reasoning. Overall, SpatialForge-2B achieves 50.6 overall average, a +8.0 absolute improvement over Qwen3-VL-2B (42.6).

\subsection{Qualitative Examples on MindCube}

\begin{figure}[t]
  \centering
  \includegraphics[width=0.9\columnwidth]{ 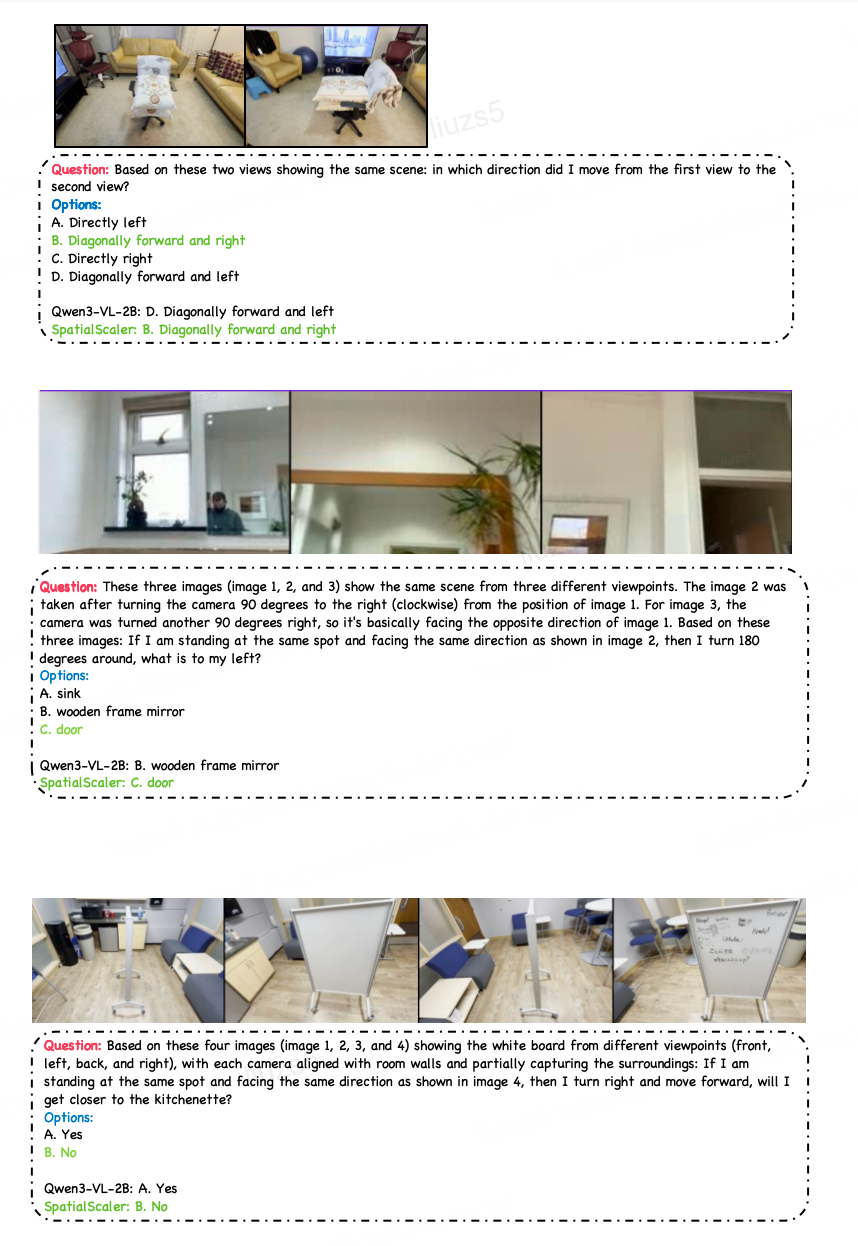}
  \caption{Visualization of Results on MindCube}
  \label{fig:mindcube}
\end{figure}

We provide qualitative examples of model predictions on MindCube in Figure~\ref{fig:mindcube}. These visualizations illustrate the model's spatial reasoning behavior in tasks such as mental rotation and perspective taking.

\begin{table*}[t]
\centering
\caption{\textbf{Detailed performance breakdown on SPAR-Bench~\citep{zhang2025flatland}.} The best results are highlighted in \textbf{bold} and the second-best results are underlined. OO, OC, and MV refer to object-object, object-camera, and multi-view, respectively. }
\label{tab:detailed_results}
\setlength{\tabcolsep}{1.5pt}
\tiny
\begin{tabular}{l|c|ccccccccc|cccc|ccccccccccc}
\toprule

Method & \rotatebox{90}{\textbf{Avg.}} & \rotatebox{90}{\textbf{Low}} & \rotatebox{90}{Depth-OC} & \rotatebox{90}{Depth-OC-MV} & \rotatebox{90}{Depth-OO} & \rotatebox{90}{Depth-OO-MV} & \rotatebox{90}{Dist-OC} & \rotatebox{90}{Dist-OC-MV} & \rotatebox{90}{Dist-OO} & \rotatebox{90}{Dist-OO-MV}  & \rotatebox{90}{\textbf{Medium}} & \rotatebox{90}{PosMatch} & \rotatebox{90}{CamMotion} & \rotatebox{90}{ViewChgI} & \rotatebox{90}{\textbf{High}} & \rotatebox{90}{DistI-OO} & \rotatebox{90}{DistI-OO-MV} & \rotatebox{90}{ObjRel-OC-MV} & \rotatebox{90}{ObjRel-OO} & \rotatebox{90}{ObjRel-OO-MV} & \rotatebox{90}{SpImag-OC} & \rotatebox{90}{SpImag-OC-MV} & \rotatebox{90}{SpImag-OO} & \rotatebox{90}{SpImag-OO-MV} \\
\midrule
\rowcolor{gray!10} \multicolumn{24}{l}{\textit{Baseline}} \\
Chance Level (Random)  & - & - & - & - & - & - & - & - & - & - & - & 22.7 & 24.5 & - & 25.1 & 23.8 & 22.0 & 31.3 & 25.3 & 22.2 & 25.8 & 24.4 & 24.2 & 26.9 \\
Chance Level (Frequency) & 32.7 & 31.2 & 43.1 & 43.5 & 17.4 & 13.1 & 41.9 & 31.0 & 27.4 & 32.2 & 38.3 & 29.0 & 26.8 & 59.0 & 32.3 & 52.9 & 50.6 & 28.3 & 26.9 & 26.6 & 26.3 & 26.7 & 26.5 & 25.8 \\
\midrule
\rowcolor{gray!10} \multicolumn{24}{l}{\textit{Open-Source Models}} \\
LLava-Onevision-7B~\citep{li2024llava} & 31.2 & 21.8 & 30.3 & 26.9 & 18.6 & 13.9 & 10.4 & 13.6 & 31.2 & 29.3 & 26.1 & \underline{38.7} & \textbf{30.3} & 9.5 & 40.1 & 56.5 & \underline{55.1} & 37.3 & 48.6 & 38.2 & 30.4 & \underline{33.7} & 26.5 & \textbf{35.0} \\
Qwen2.5-VL-3B~\citep{yang2025qwen3} & 24.6 & 19.4 & 38.0 & 40.6 & 18.8 & 14.1 & 7.8 & 7.1 & 17.8 & 11.1 & 27.6 & 26.2 & 25.3 & \textbf{31.2} & 28.2 & 54.1 & 49.1 & 21.8 & 25.3 & 12.5 & 23.9 & 27.6 & 24.8 & 14.9\\
Qwen2.5-VL-7B~\citep{yang2025qwen3} & 33.1 & 28.8 & 31.3 & 33.7 & 22.0 & 15.0 & 42.9 & 37.7 & 23.8 & 23.6 & 23.0 & 33.3 & \underline{28.8} & 6.8 & 40.3 & \underline{58.2} & 51.5 & 44.8 & \underline{50.0} & 32.1 & \underline{33.9} & 32.9 & \underline{27.2} & 31.9 \\
Qwen3-VL-2B~\cite{yang2025qwen3} & \underline{42.6} & \underline{60.6} & \textbf{59.2} & \textbf{58.3} & \underline{51.9} & \underline{49.2} & \underline{66.3} & \textbf{68.3} & \underline{68.6} & \underline{67.7} & \underline{27.4} & 2.5 & 23.0 & 16.4 & \underline{41.2} & 57.1 & 54.2 & \underline{45.5} & 45.6 & \underline{38.5} & 27.2 & 33.1 & 26.5 & 30.3\\
\midrule
\rowcolor{gray!20} \textbf{SpatialForge-2B (Ours)} 
& \textbf{50.6} 
& \textbf{65.8} 
& \underline{54.8} 
& \underline{56.5} 
& \textbf{68.2} 
& \textbf{68.5} 
& \textbf{67.7} 
& \underline{65.5} 
& \textbf{73.1} 
& \textbf{72.3} 
& \textbf{32.3} 
& \textbf{49.9} 
& 28.0
& \underline{19.1}
& \textbf{43.6} 
& \textbf{60.9} 
& \textbf{58.6} 
& \textbf{48.3} 
& \textbf{53.6} 
& \textbf{39.6} 
& \textbf{36.1} 
& \textbf{36.1} 
& \textbf{29.5} 
& \underline{32.5} \\
\bottomrule
\end{tabular}
\end{table*}

\section{License}
\label{appendix:license}
We conduct a systematic review of the open-source licenses for the datasets used in our data construction pipeline, with the results summarized in Table~\ref{tab:licenses}. Due to the use of multi-source data, our dataset inherits a variety of licenses, which are listed accordingly in the table.
\begin{table*}[h]
\centering
\caption{\textbf{The licenses for the datasets and benchmarks included in this paper.}}
\footnotesize
\label{tab:licenses}
\renewcommand{\arraystretch}{1.15}
\setlength{\tabcolsep}{6pt}

\begin{tabular}{@{}lll@{}}
\toprule
\textbf{Dataset} & \textbf{Type} & \textbf{License} \\
\midrule

\multicolumn{3}{c}{\textit{\textbf{Benchmarks}}} \\

CV-Bench (2D \& 3D)~\citep{tong2024cambrian}
& Indoor, Outdoor
& Apache License 2.0 \\

SPAR-Bench~\citep{zhang2025flatland}
& Indoor
& MIT License \\

SpaCE10~\citep{Gong2025SpaCE10AC}
& Indoor
& MIT License \\

OmniSpatial~\citep{jia2025omnispatial}
& General spatial scenes
& Apache License 2.0 \\

MindCube~\citep{yin2025spatial}
& Multi-view spatial scenes
& MIT License \\

\midrule

\multicolumn{3}{c}{\textit{\textbf{Source Datasets}}} \\

Objects365~\citep{shao2019objects365}
& General object-centric scenes
& CC BY 4.0  \\

OpenImages~\citep{kuznetsova2020open}
& General object-centric scenes
& CC BY 4.0  \& CC BY 2.0  \\

PixMo~\citep{deitke2025molmo}
& General scenes
& ODC-BY-1.0 \\

SpatialForge-10M (Ours)
& General scenes
& CC BY 4.0 \& CC BY 2.0 \& ODC-BY-1.0 \\
\bottomrule
\end{tabular}
\end{table*}

\section{Broader Impacts and Safeguards}
\label{appendix:impact}
This paper focuses on advancing fundamental spatial reasoning capabilities in Vision-Language Models (VLMs) through a synthetic data pipeline. The primary contribution is the construction of a large-scale dataset and benchmark for spatial reasoning. Potential positive societal impacts include applications in robotic navigation, assistive technologies for visually impaired individuals, autonomous driving, and augmented reality systems.
The proposed pipeline poses minimal safety and privacy risks. All supervision signals contain only geometric and positional information. The dataset does not include personally identifiable information, sensitive attributes, or unsafe content. In addition, our work focuses on spatial understanding rather than image generation, and therefore does not introduce risks related to deepfakes, impersonation, or misinformation.
To further ensure safety, the dataset is constructed through an automated synthesis process using controlled and filtered data sources, avoiding offensive or harmful content. Overall, we believe the societal risk of misuse is low.
\section{Prompt Used in Data Synthesis Pipeline}
\label{appendix:prompt for data_preprocess}
\tcbset{
    stagebox/.style={
        colback=white,
        colframe=black,
        fonttitle=\bfseries,
        coltitle=white,
        colbacktitle=black,
        boxrule=0.5pt,
        arc=2mm,
        outer arc=2mm,
        shadow={0.5mm}{-0.5mm}{0.5mm}{black!50!white},
        left=2mm,
        right=2mm,
        top=2mm,
        bottom=2mm,
        enhanced,
    }
}
\begin{tcolorbox}[stagebox,breakable,title=Prompt for Global Captioning and Object Category Extraction]
\textbf{System Prompt:} ``You are a helpful vision assistant specialized in detailed image understanding and object category extraction. \\
You will be shown an image. \\
Your task is to analyze the entire image and produce two outputs: \\
(1) a comprehensive, faithful, information-dense caption of the entire image; \\
(2) a Python list of all visible object categories appearing in the image. \\[1mm]

When generating the caption, follow these requirements: \\
- Explicitly name as many salient visible objects as possible using concrete nouns. \\
- Mention important attributes such as color, material, size, state, or appearance when visible. \\
- Describe relative positions and layout, such as left/right/center, front/back, on/under/next to. \\
- Include actions or interactions if they are clearly visible. \\
- Mention background and environment only if they help explain the scene. \\
- Avoid vague placeholders such as `object', `item', `stuff', or `something'. \\
- Generate a faithful, information-dense caption in 3--4 sentences. \\[1mm]

When extracting object categories, follow these requirements: \\
- Extract only core object names that are part of the scene depicted in the image. \\
- Exclude modifiers and descriptive attributes, such as color, size, material, or state. \\
- Do not extract meta-terms about the image format, such as `image', `photo', `picture', `portrait', `photograph', `illustration', `painting', `drawing', `sketch', or `artwork'. \\
- For real-world scenes, extract physical elements and exclude pure backgrounds such as sky, wall, floor, or surface unless they are semantically important. \\
- For posters, graphics, artworks, or paintings, extract the compositional elements. \\
- For screenshots or interfaces, extract user interface elements such as buttons, icons, menus, and text boxes. \\
- Each object category should be a singular, generic noun, for example, use `car' instead of `yellow taxi'. \\
- Merge similar object categories, for example, combine `building' and `skyscraper' as `building'. \\[1mm]

\textbf{Output Format:} \\
Return the result using exactly the following format: \\
\texttt{Caption: <3--4 sentence detailed caption>} \\
\texttt{Objects: ["object1", "object2", "object3"]} \\
Do not include any additional comments or explanations.'' \\[1mm]

\textbf{User Prompt:} \{question\} + ``Please analyze the entire image. First, provide a detailed description of the whole image. Then, extract all visible object categories from the description and return them as a Python list of strings.''
\end{tcolorbox}

\begin{tcolorbox}[stagebox,breakable,title=Prompt for Region Captioner]
\textbf{System Prompt:} ``You are a careful vision inspector. \\
You will see an image and one target bounding box. \\
Only judge the visual content strictly inside the bounding box. \\[1mm]

\textbf{Task:} Based on the object hint and the visual content inside the bounding box, write a brief description of what is inside the bounding box. \\[1mm]

\textbf{Person Region:} \\
If the object hint refers to a person, such as man, woman, child, person, or people, describe the visible content in the following order, skipping any invisible attributes: \\
1. Gender or age group, such as man, woman, elderly man, young girl, or boy. \\
2. Position in the image, using spatial terms such as left, right, center, front, back, top, bottom, foreground, or background. \\
3. Facing direction, such as facing camera, facing left, back to camera, or profile view. \\
4. Upper body clothing, including color and garment type, such as white shirt, red hoodie, or blue suit jacket. \\
5. Lower body clothing if visible, such as black jeans or grey skirt. \\
6. One prominent accessory or feature if notable, such as hat, glasses, backpack, or long hair. \\[1mm]

Example outputs: \\
``Woman facing camera wearing red blouse and blue jeans, sunglasses in the foreground'' \\
``Elderly man back to camera in grey coat and dark trousers, on the left'' \\
``Young boy in the background, in yellow t-shirt, facing left'' \\[1mm]

\textbf{Object Region:} \\
If the object hint refers to a non-person thing, describe the visible content in the following order, skipping any invisible attributes: \\
1. Object name, using the provided object hint. \\
2. Position in the image, using spatial terms such as left, right, center, front, back, top, bottom, foreground, or background. \\
3. Dominant color or visible pattern. \\
4. Material, such as metal, wood, plastic, fabric, glass, or ceramic, when distinguishable. \\
5. Shape, size, or quantity if notable, such as long, round, small, or a pair of. \\
6. One distinctive feature or state, such as open, broken, stacked, or worn. \\[1mm]

Example outputs: \\
``Red plastic bottle with white screw cap, cylindrical, located on the right side'' \\
``Wooden chair with blue fabric cushion, four legs, positioned in the center foreground'' \\
``Silver metal fork, long thin handle, placed on the left side of the image'' \\
``Stack of white ceramic plates, smooth surface, located in the background on the left'' \\[1mm]

\textbf{Position Description Examples:} \\
- Spatial: on the left/right side, in the center, at the top/bottom. \\
- Depth: in the foreground, in the background, in the middle ground. \\[1mm]

\textbf{Rules:} \\
- Only describe what is visible inside the bounding box; ignore everything outside it. \\
- Start with the most identifying term: gender or age group for persons, object name for things. \\
- Use the object hint as the starting object name when it refers to a non-person thing. \\
- Keep the description brief, under 20 words total. \\
- Do not include extra formatting, explanations, or punctuation at the end. \\
- Output only the description.'' \\[1mm]

\textbf{User Prompt:} \\
\{region\_note\} \\
\{fisheye\_note\} \\
Object category: \{hint\}. You MUST start your description based on this name. \\
\{position\_hint\} \\
Target bbox \texttt{(xmin,ymin,xmax,ymax)}: \texttt{(\{xmin\},\{ymin\},\{xmax\},\{ymax\})} on image size \texttt{(W=\{W\}, H=\{H\})}. \\
Provide a brief description starting with the object name, then its key visual attributes under 20 words total.
\end{tcolorbox}

\begin{tcolorbox}[stagebox,breakable,title=Prompt for Orientation Estimation]
\textbf{System Prompt:} ``You are a detailed vision labeler. \\
You will see an image and one target bounding box. \\
Only describe the visual content inside that bounding box. \\[1mm]

\textbf{Task:} Output a JSON object with exactly two keys: \\
1. \texttt{description}: a simple noun phrase describing the person, up to 10 words. \\
2. \texttt{facing}: the direction the person is facing. \\[1mm]

\textbf{Description Rules:} \\
- Start with \texttt{the} plus an identity word, such as man, woman, boy, girl, person, athlete, etc. \\
- Describe visible clothing features. \\
- Include age cue, gender if clear, and dominant colors when visible. \\
- Or include notable accessories, such as glasses, hat, bag, helmet, uniform number, etc. \\
- Or include action or pose if clear, such as standing, sitting, running, jumping, or crouching. \\
- Do not include location or position words, such as left, right, center, foreground, or background. \\
- Do not use punctuation at the end, full sentences, or conjunctions such as \texttt{who}. \\[1mm]

\textbf{Facing Values:} Choose exactly one value from the following: \\
- \texttt{front}: face clearly visible, looking toward camera. \\
- \texttt{back}: back of head or body visible, facing away. \\
- \texttt{left}: person faces to their left. \\
- \texttt{right}: person faces to their right. \\
- \texttt{side}: profile view, either left or right. \\
- \texttt{three-quarter}: angled, not fully front or side. \\
- \texttt{unknown}: cannot determine. \\[1mm]

\textbf{Output Format:} \\
Output only valid JSON, with no markdown and no extra text. \\[1mm]

Examples: \\
\texttt{\{"description": "the young man in white hockey helmet and black jersey", "facing": "front"\}} \\
\texttt{\{"description": "the elderly woman in gray cardigan and floral skirt and brown loafers", "facing": "back"\}} \\
\texttt{\{"description": "the teenage boy in blue baseball cap and yellow hoodie", "facing": "side"\}}'' \\[2mm]

\end{tcolorbox}




\section{Limitations.}
\label{appendix:limitation}
Despite its effectiveness, our approach has several limitations. 
First, inferring 3D spatial relationships from 2D images is inherently ambiguous, and the resulting supervision is approximate, especially in scenarios involving occlusion, perspective distortion, or complex scene layouts. 

Second, our multi-stage data construction pipeline (e.g., captioning, detection, and geometric estimation) may introduce error accumulation, despite the use of verification mechanisms. 
In particular, the reliance on monocular depth estimation can lead to unreliable signals in challenging cases such as reflective surfaces or thin structures.

Third, our framework focuses on a limited set of spatial relations, such as depth ordering and horizontal layout, and does not yet cover more complex spatial reasoning involving physical interactions or temporal dynamics. 

Fourth, we also observe a mild trade-off between spatial reasoning and general perception: while SpatialForge significantly improves reasoning performance, it may slightly affect performance on purely 2D perception tasks due to model capacity being reallocated toward geometric understanding.
Finally, compared to approaches that leverage explicit 3D data, our method does not provide precise metric geometry, which may limit performance on tasks requiring fine-grained spatial accuracy. 
Addressing these limitations by improving supervision quality and expanding task coverage is an important direction for future work.


\end{document}